\documentclass[sigconf]{acmart}
\AtBeginDocument{%
  }

\usepackage{xspace}
\usepackage{array}
\usepackage{tabularx}
\usepackage{multirow}
\usepackage{booktabs}
\usepackage{graphicx}
\usepackage{amsmath}
\usepackage{hyperref}
\usepackage{pifont}
\usepackage{algorithm}
\usepackage{algpseudocode}

\usepackage{marvosym}

\setcopyright{acmlicensed}
\copyrightyear{2018}
\acmYear{2018}
\acmDOI{XXXXXXX.XXXXXXX}
\acmConference[Conference acronym 'XX]{Make sure to enter the correct
  conference title from your rights confirmation email}{June 03--05,
  2018}{Woodstock, NY}
\acmISBN{978-1-4503-XXXX-X/2018/06}

\newcommand{\PMcoarse}{$\mathrm{PM}_{10}$\xspace}
\newcommand{\PM}{$\mathrm{PM}_{2.5}$\xspace}
\newcommand{\ozone}{$\mathrm{O}_{3}$\xspace}
\newcommand{\OurModel}{PCDCNet\xspace}

\newcommand{\Ammonia}{$\mathrm{NH}_3$\xspace}
\newcommand{\NitrogenOxides}{$\mathrm{NO}_x$\xspace}
\newcommand{\SulfurDioxide}{$\mathrm{SO}_2$\xspace}
\newcommand{\VOC}{$\mathrm{VOC}$\xspace}
\newcommand{\AirPoll}{$\mathbf{X}$\xspace}
\newcommand{\Meteo}{$\mathbf{P}$\xspace}
\newcommand{\Emiss}{$\mathbf{Q}$\xspace}
\newcommand{\cmark}{\ding{51}}%
\newcommand{\xmark}{\ding{55}}%
\begin{document}


\title{PCDCNet: A Surrogate Model for Air Quality Forecasting with Physical-Chemical Dynamics and Constraints}


\author{Shuo Wang}
\email{shuowang.ai@gmail.com}
\affiliation{
  \institution{
  School of Systems Science\\
  Beijing Normal University}
  \city{Beijing}
  \country{China}
}
\affiliation{
  \institution{D-ITET, ETH Zurich}
  \city{Zurich}
  \country{Switzerland}
}

\author{Yun Cheng}
\email{yun.cheng@sdsc.ethz.ch}
\affiliation{
  \institution{Swiss Data Science Center\\ETH Zurich}
  \city{Zurich}
  \country{Switzerland}
}

\author{Qingye Meng}
\email{hilbertmeng@gmail.com}
\affiliation{
  \institution{ColorfulClouds\\Technology Co.,Ltd.}
  \city{Beijing}
  \country{China}
}

\author{Olga Saukh}
\email{saukh@tugraz.at}
\affiliation{
  \institution{Graz University of Technology}
  \city{Graz}
  \country{Austria}
}
\affiliation{
  \institution{Complexity Science Hub}
  \city{Vienna}
  \country{Austria}
}

\author{Jiang Zhang}
\email{zhangjiang@bnu.edu.cn}
\affiliation{
  \institution{School of Systems Science\\Beijing Normal University}
  \city{Beijing}
  \country{China}
}

\author{Jingfang Fan}
\email{jingfang@bnu.edu.cn}
\authornote{Corresponding author.}
\affiliation{
  \institution{School of Systems Science / Institute of Nonequilibrium Systems\\Beijing Normal University}
  \city{Beijing}
  \country{China}
}
\affiliation{
  \institution{Potsdam Institute for Climate Impact Research}
  \city{Potsdam}
  \country{Germany}
}

\author{Yuanting Zhang}
\email{zhangyuanting@caiyunapp.com}
\affiliation{
  \institution{ColorfulClouds\\Technology Co.,Ltd.}
  \city{Beijing}
  \country{China}
}

\author{Xingyuan Yuan}
\email{yuan@caiyunapp.com}
\affiliation{
  \institution{ColorfulClouds\\Technology Co.,Ltd.}
  \city{Beijing}
  \country{China}
}

\author{Lothar Thiele}
\email{thiele@ethz.ch}
\affiliation{
  \institution{D-ITET, ETH Zurich}
  \city{Zurich}
  \country{Switzerland}
}

\renewcommand{\shortauthors}{Shuo Wang et al.}

\begin{abstract}
Air quality forecasting (AQF) is critical for public health and environmental management, yet remains challenging due to the complex interplay of emissions, meteorology, and chemical transformations. Traditional numerical models, such as CMAQ and WRF-Chem, provide physically grounded simulations but are computationally expensive and rely on uncertain emission inventories. Deep learning models, while computationally efficient, often struggle with generalization due to their lack of physical constraints. To bridge this gap, we propose \textbf{\OurModel}, a \textbf{surrogate model} that integrates numerical modeling principles with deep learning. \OurModel explicitly incorporates emissions, meteorological influences, and domain-informed constraints to model pollutant formation, transport, and dissipation. By combining \textit{graph-based spatial transport modeling}, \textit{recurrent structures for temporal accumulation}, and \textit{representation enhancement for local interactions}, \OurModel achieves \textbf{state-of-the-art (SOTA)} performance in \textbf{72-hour station-level} \PM and \ozone forecasting while significantly reducing computational costs. Furthermore, our model is deployed in an online platform, providing free, real-time air quality forecasts, demonstrating its scalability and societal impact. By aligning deep learning with \textbf{physical consistency}, \OurModel offers a practical and interpretable solution for AQF, enabling informed decision-making for both personal and regulatory applications.
\end{abstract}

\ccsdesc[500]{Computing methodologies~Neural networks}
\ccsdesc[100]{Applied computing~Physics}
\ccsdesc[300]{Applied computing~Environmental sciences}

\keywords{Air quality forecasting, physical-chemical dynamics, PINNs, meteorology, emissions, deep learning, graph neural networks, surrogate modeling, photochemical reactions, real-time forecasting.}

\received{20 February 2007}
\received[revised]{12 March 2009}
\received[accepted]{5 June 2009}

\maketitle

\section{Introduction}

\begin{figure}[t]
  \centering
  \includegraphics[width=\linewidth]{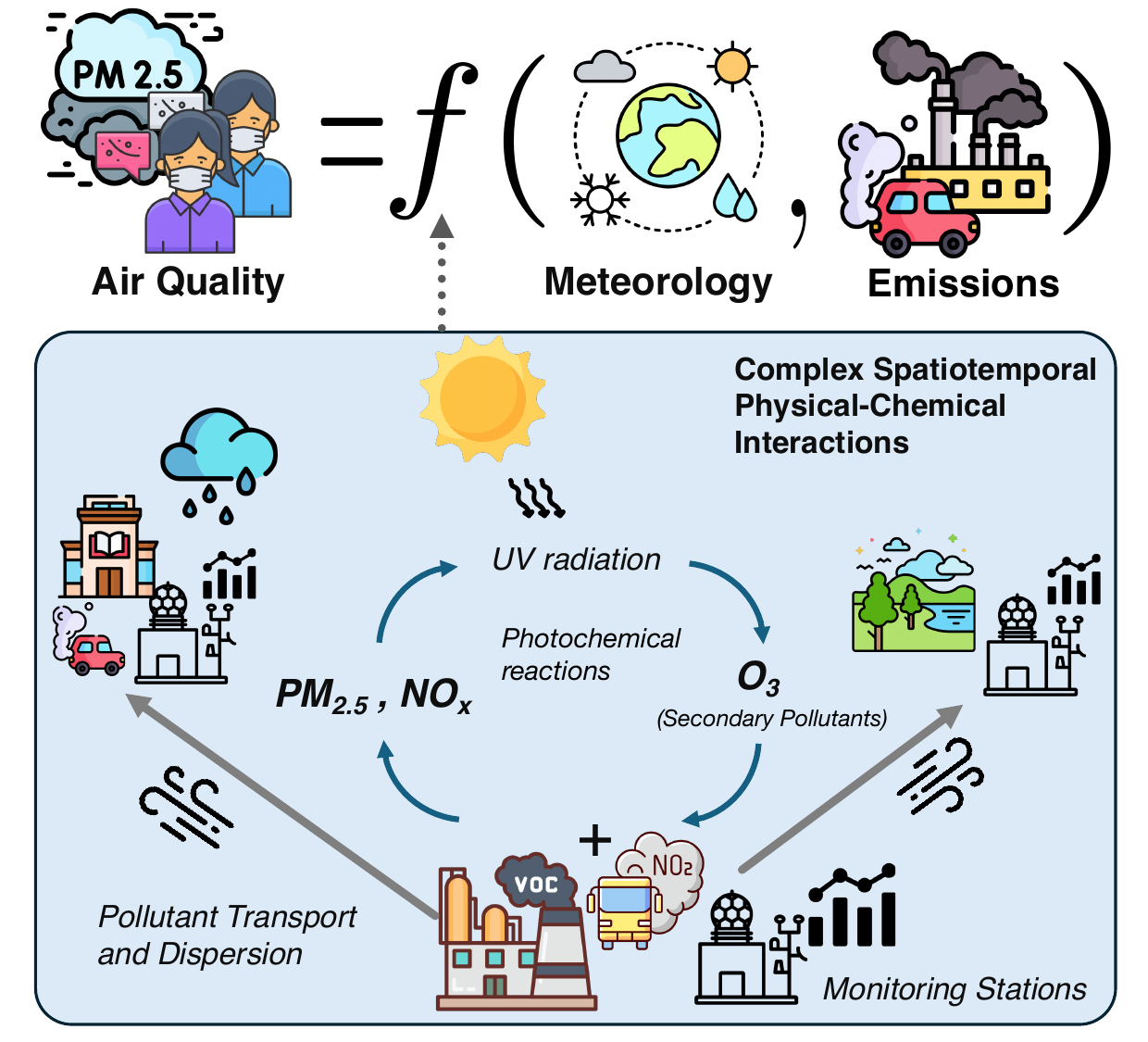}
  \caption{Air quality (\PM and \ozone) is shaped by complex interactions between meteorology (e.g., UV radiation, wind) and emissions (e.g., \NitrogenOxides, \VOC). Capturing these spatiotemporal dynamics, including pollutant transport and secondary formation, requires integrating emissions data with meteorology, which poses significant modeling challenges.}
  \Description{}
  \label{fig:intro}
\end{figure}

Air pollution remains a pressing global challenge, disproportionately affecting developing nations such as China, India, Pakistan, and Bangladesh~\cite{brauer2016ambient, yang2023socio, anwar2021emerging}. Over the years, China has implemented rigorous measures to improve air quality, achieving significant reductions in \PM concentrations~\cite{geng2024efficacy}. However, substantial gaps persist compared to the air quality standards of developed countries~\cite{yue2020stronger}. Furthermore, the reduction in \PM levels has inadvertently heightened the prominence of \ozone pollution~\cite{lu2020rapid,einsiedler2021interpretabletransferablemodelsunderstand}, exposing the intricate and nonlinear interactions between these pollutants and emphasizing the need for coordinated air pollution management strategies~\cite{li2019two, chen2024region}.

In developed nations, while routine air quality levels are relatively better, catastrophic events like wildfires in Australia, the United States, and Canada have triggered severe pollution episodes, endangering public health~\cite{chen2025canadian, burke2023contribution, shrestha2022observations}. These scenarios underscore the necessity of real-time, high-precision air quality forecasting (AQF) systems that can provide reliable site-specific predictions up to three days in advance. Such systems can serve as critical tools for personal protection, travel planning, and government interventions aimed at regulating industrial emissions, ultimately contributing to public health and environmental sustainability.

Air quality forecasting, particularly for key pollutants like \PM and \ozone, is inherently challenging due to the complex interplay of emissions, meteorology, and secondary chemical reactions~\cite{xiao2021tracking, wang2019responses, sun2022meteorological}. As illustrated in Figure~\ref{fig:intro}, pollutant concentrations are influenced by long-range spatiotemporal dependencies, where pollutants can travel vast distances under the influence of wind and atmospheric dynamics~\cite{chen2020influence, chen2018understanding}. This results in local air quality being shaped not only by nearby emissions but also by regional and even global factors. Compounding this complexity, spatially proximate regions can exhibit starkly different pollution patterns due to variations in emission sources and industrial structures~\cite{li2017anthropogenic, zheng2018trends, geng2024efficacy}. Moreover, many pollutants, such as \ozone, undergo complex chemical transformations, being formed as secondary pollutants through photochemical reactions involving \NitrogenOxides\ and \VOC\ under sunlight \cite{qu2023underlying, li2024exploring}. These challenges necessitate the development of advanced models capable of integrating emissions and meteorology, capturing long-range dependencies, and modeling pollutant interactions.

Traditional numerical models like CMAQ (Community Multiscale Air Quality model)~\cite{park2021implementation, li2022evaluation}\footnote{\url{https://www.epa.gov/cmaq}} and WRF-Chem~\cite{sicard2021high}\footnote{\url{https://www2.acom.ucar.edu/wrf-chem}} have long been the cornerstone of AQF~\cite{gao2024review}. These models solve partial differential equations (PDEs) to simulate the physical and chemical processes underlying pollutant formation, transport, and transformation. While these approaches provide interpretable insights, they are computationally expensive and unsuitable for real-time forecasting.

Moreover, their accuracy heavily depends on precisely estimated initial conditions, which are often difficult to obtain in practice~\cite{hou2022impacts,rao2020limit}. This reliance introduces significant uncertainty, especially in dynamic scenarios such as sudden pollution events. Additionally, these models are typically used retrospectively and lack native integration of real-time observational data~\cite{li2022evaluation}, further limiting their adaptability to rapidly changing environmental conditions.

The advent of AI-based Earth system models, such as Pangu-Weather~\cite{bi2023accurate}, GraphCast~\cite{lam2023learning}, and Aurora~\cite{bodnar2024aurora}, have introduced a paradigm shift in geosciences, leveraging the strong pattern recognition capabilities of AI to efficiently model complex atmospheric phenomena. However, these models primarily focus on meteorological forecasting and often neglect the intricacies of air quality data. They typically operate on grid-based datasets like ERA5 reanalysis, which lack the spatial granularity and station-level variability of air quality monitoring systems.

Recently proposed transformer-based models, such as iTransformer~\cite{liu2023itransformer} and TimeXer~\cite{wang2024timexer}, have demonstrated remarkable advances on long-term time-series forecasting tasks. Yet, their generalized architectures are not specifically tailored for AQF. These models often fail to meet the high accuracy demands of AQF because they focus on learning generic temporal patterns rather than domain-specific dynamics like pollutant transport, interactions, and secondary formation. Existing AQF methods, on the other hand, either exclude critical variables like emissions data~\cite{wang2020pm25} or fail to adequately capture physical and chemical mechanisms~\cite{wang2021modeling, hettige2024airphynet}, leading to limited generalizability and suboptimal performance.

To this end, we introduce \textbf{\OurModel}, a \textbf{surrogate model for AQF} that bridges numerical simulations and deep learning. \OurModel captures \textbf{physical-chemical dynamics} while maintaining computational efficiency, serving as a data-driven alternative to traditional models. Unlike black-box deep learning methods, it explicitly integrates emissions, meteorology, and domain-informed constraints, ensuring both accuracy and interpretability. The key contributions of this work are as follows:

\begin{itemize}
    \item \textbf{We propose \OurModel, a surrogate modeling framework for AQF that ensures physical consistency while aligning with CMAQ-like numerical simulations.} By integrating emissions, meteorology, and spatiotemporal dynamics within a structured input-output framework, \OurModel enforces atmospheric constraints, enabling interpretable and robust predictions.

    \item \textbf{We develop a hybrid architecture incorporating physical and chemical dynamics into deep learning, ensuring consistency and generalization.} \OurModel leverages GRU for temporal accumulation, graph-based spatial transport modeling, and representation learning for pollutant interactions, with domain-informed constraints enforcing alignment with physical laws.

    \item \textbf{We deploy \OurModel as a real-time AQF service, offering free and accessible forecasts.} Integrated into an online platform\footnote{\href{https://caiyunapp.com/map/\#116.3507,40.0099}{\OurModel-powered AQF: https://caiyunapp.com/map}}, it delivers 72-hour station-level predictions, providing actionable insights for health protection, travel planning, and policy decisions.
\end{itemize}

By introducing a \textbf{physically consistent and computationally efficient} surrogate modeling paradigm, \OurModel bridges numerical simulations and deep learning, reinforcing the role of domain knowledge in AI-driven environmental modeling.

\section{Preliminaries}

This section presents the datasets, research regions, spatial graph construction, and problem definition for AQF. Additionally, we compare input-output paradigms across numerical models, deep learning methods, and our approach.

\subsection{Datasets and Research Regions}

Our study utilizes air quality, meteorological, and emission datasets to tackle AQF challenges in two key regions of China: the \textbf{Beijing-Tianjin-Hebei and Surrounding Areas (BTHSA)}, spanning approximately \textbf{430,000 km²} and covering \textbf{2+26 cities}~\cite{lu2021estimation}\footnote{\url{https://www.mee.gov.cn/gkml/sthjbgw/qt/201701/t20170124_395229.htm}}, and the \textbf{Yangtze River Delta (YRD)}, encompassing around \textbf{270,000 km²}. These regions exhibit diverse pollution characteristics, meteorological variations, and emission patterns, providing a comprehensive testbed for evaluating AQF models, as depicted in Figure~\ref{fig:regions_graph}. Appendix~\ref{appendix:variables} summarizes the variables used in this study.

\paragraph{Air Quality Data.} Hourly \PM and \ozone concentrations from 355 monitoring stations across \textbf{BTHSA} and \textbf{YRD} (Figure~\ref{fig:regions_graph}) were sourced from CNEMC\footnote{\url{https://www.cnemc.cn/}}. These datasets provide high spatial and temporal resolution, enabling detailed analysis of pollution dynamics in both urban and rural areas.

\paragraph{Meteorological Data.} Key meteorological variables, including wind components (u100, v100), temperature (t2m), precipitation (tp), surface pressure (sp), and others, were sourced from the ERA5 reanalysis dataset\footnote{\url{https://cds.climate.copernicus.eu/datasets/reanalysis-era5-single-levels}}. For deployment, we utilized GFS forecast data\footnote{\url{https://www.ncei.noaa.gov/products/weather-climate-models/global-forecast}}, which provides hourly predictions updated four times daily.

\paragraph{Emission Data.} Regional emissions of \NitrogenOxides, \VOC, \SulfurDioxide, \Ammonia, and \PM were obtained from the Multi-resolution Emission Inventory for China (MEIC) \cite{li2017anthropogenic, geng2024efficacy}\footnote{\url{http://meicmodel.org.cn/}}. These monthly inventories, while statistical estimates, capture general emission trends and were downscaled to an hourly resolution using the methodology in~\cite{inventory} to align with air quality and meteorological data.

\paragraph{Graph Construction.} A spatial adjacency graph was constructed by connecting monitoring stations within a 200km geodesic threshold, as shown in Figure~\ref{fig:regions_graph}, following~\cite{qi2019hybrid, wang2021modeling}. This graph models pollutant transport pathways and underpins the GNN-based components of our framework.

\subsection{Problem Statement, Inputs and Outputs}

The goal of AQF is to predict future pollutant concentrations using historical data and forecasted meteorology and emissions. As shown in Figure~\ref{fig:input_output}, given historical pollutant concentrations $\mathbf{X}^{-T^{\prime}+1:0}$, meteorological variables $\mathbf{P}^{-T^{\prime}+1:0}$, and emissions $\mathbf{Q}^{-T^{\prime}+1:0}$, along with forecasted meteorology $\mathbf{P}^{1:T}$ and emissions $\mathbf{Q}^{1:T}$, the objective is to estimate $\hat{\mathbf{X}}^{1:T}$:

\begin{equation}
\hat{\mathbf{X}}^{1:T} = \mathcal{F}_\Theta \left(\mathbf{X}^{-T^{\prime}+1:0}, \mathbf{P}^{-T^{\prime}+1:T}, \mathbf{Q}^{-T^{\prime}+1:T} \right),
\label{eq:problem_formulation}
\end{equation}

\noindent where $\mathcal{F}_\Theta$ models spatiotemporal dependencies, pollutant transport, and secondary pollutant formation.

\section{Methodology}

\begin{figure*}[h]
  \centering
  \includegraphics[width=\linewidth]{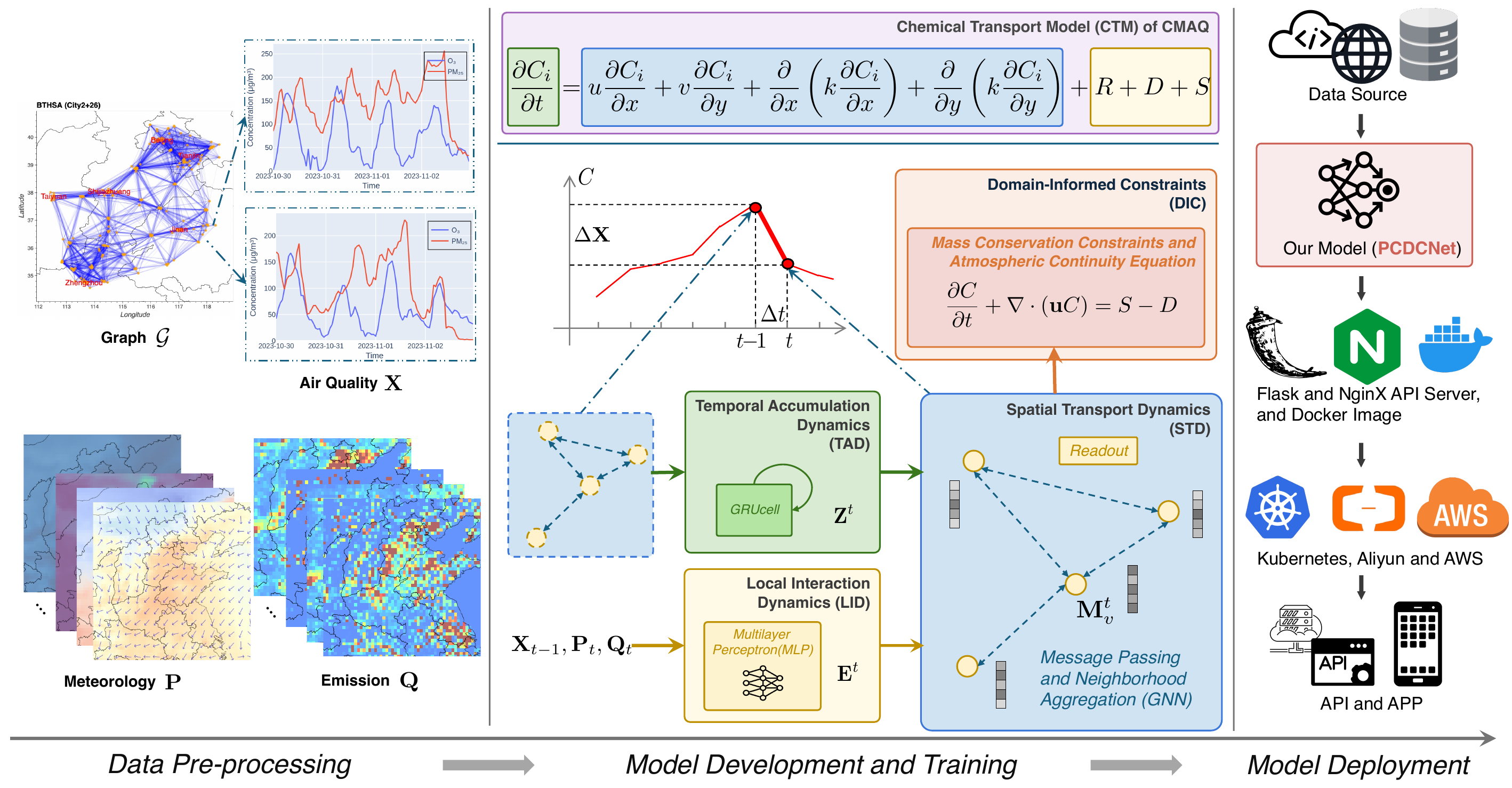}
  \caption{The framework of \OurModel for air quality forecasting (AQF), comprising three stages: Data Pre-processing (graph construction and integration of $\mathbf{X}$, $\mathbf{P}$, $\mathbf{Q}$), Model Development (modules for temporal, spatial, and local dynamics with domain-informed constraints), and Model Deployment (real-time predictions via cloud-based Dockerized APIs).}
  \Description{}
  \label{fig:overall_framework}
\end{figure*}

This section introduces \OurModel, a surrogate model that integrates emissions, meteorology, and physical-chemical constraints for efficient and interpretable AQF. By embedding atmospheric processes such as \textbf{advection, diffusion, deposition, and secondary formation}, \OurModel bridges the gap between \textbf{numerical simulations (e.g., CMAQ)} and \textbf{AI-based methods}.

\subsection{Model Overview}

Traditional chemical transport models (CTMs), such as CMAQ, describe pollutant evolution through a set of partial differential equations (PDEs) that account for various atmospheric processes, including advection, diffusion, deposition, emissions, and chemical reactions~\cite{hobbs2000introduction, seinfeld2016atmospheric}. The general form of pollutant concentration $C$ at time $t$ is governed by:

\begin{equation}
\frac{\partial C_i}{\partial t} = -\mathbf{u} \cdot \nabla C_i + \nabla \cdot (k \nabla C_i) + R + D + S,
\label{eq:ctm}
\end{equation}

where $\mathbf{u}$ represents wind vectors (advection), $k$ is the diffusion coefficient, $R$ accounts for chemical reactions, and $D$, $S$ denote deposition and source terms, respectively.

Traditional methods face significant limitations, including their dependence on preprocessed meteorological inputs (e.g., WRF), which introduces uncertainties and restricts the use of direct NWP products like GFS and ECMWF, as well as reliance on initial condition fields with high uncertainty impacting prediction accuracy. Additionally, limited integration with station-level data reduces adaptability to local pollution dynamics, while high computational costs hinder their feasibility for real-time applications.

To overcome traditional limitations, we introduce \textbf{\OurModel}, a deep learning-based surrogate model for air quality forecasting (AQF) that integrates the physical and chemical principles of CTMs with data-driven learning. As shown in Figure~\ref{fig:overall_framework}, \OurModel incorporates emissions, meteorology, and pollutant dynamics within a structured spatiotemporal framework, consisting of three core modules:

\begin{itemize}
    \item \textbf{Local Interaction Dynamics (LID)}: Models localized pollutant interactions (e.g., secondary formation) using emissions, meteorology, and concentrations via an MLP.
    \item \textbf{Spatial Transport Dynamics (STD)}: Captures pollutant advection and dispersion through graph-based message passing for long-range dependencies.
    \item \textbf{Temporal Accumulation Dynamics (TAD)}: Learns temporal patterns and pollutant persistence using a GRU for historical trends and decay dynamics.
\end{itemize}

\textbf{Pseudo-code Implementation}
Algorithm~\ref{alg:algorithm} provides the pseudo-code for the training and inference process of \OurModel, highlighting the interactions among those modules. The \textbf{Embed} step maps input data into the hidden space, while the \textbf{Readout} step translates the hidden space representation into the predicted outputs.

\begin{algorithm}[h!]
\caption{Training and inference procedure of \OurModel.}
\label{alg:algorithm}
\begin{algorithmic}[1]
\Require Training dataset \( \mathcal{D} = \{D_n\}_{n=1}^N \), initialized parameters \( \Theta \), learning rate \( \alpha \)
\Ensure Optimized \( \Theta \) (training), sequence \( \{\hat{\mathbf{X}}\}_{t=1}^T \) (inference)

\For{each sample \( D_k = (\mathbf{X}, \mathbf{P}, \mathbf{Q}) \in \mathcal{D} \)}
    \State Initialize \( \mathbf{Z}^{-T'+1} = \mathbf{0} \)
    \For{\( t = -T'+2 \) to \( T \)} \Comment{\( t_0 \) marks the last observed AQI in the sample; time steps correspond to its sliding window. In online deployment, \( t_0 \) is the current time.}
        \State \( \mathbf{H}^t = 
            \begin{cases} 
                \text{Linear}([\mathbf{X}^{t-1}, \mathbf{P}^t, \mathbf{Q}^t]), & \text{if } t < 1 \\
                \text{Linear}([\hat{\mathbf{X}}^{t-1}, \mathbf{P}^t, \mathbf{Q}^t]), & \text{otherwise}
            \end{cases} \)
         \Comment{Embedding: Use ground truth \( \mathbf{X} \) for historical phase, previous estimate \( \hat{\mathbf{X}} \) for prediction.}
        \State \( \mathbf{E}^t = \text{MLP}(\mathbf{H}^t); \quad \mathbf{H}^t \mathrel{+}= \mathbf{E}^t \) \Comment{LID} 
        \State \( \mathbf{M}^t = \text{MessagePassing}(\mathbf{H}^t, \mathcal{G}); \quad \mathbf{H}^t \mathrel{+}= \mathbf{M}^t \) \Comment{STD}        
        \State \( \mathbf{Z}^t = \text{GRUcell}(\mathbf{H}^t, \mathbf{Z}^{t-1}); \quad \mathbf{H}^t \mathrel{+}= \mathbf{Z}^t \) \Comment{TAD}  
        \State \( \Delta \hat{\mathbf{X}}^{t} = \text{Linear}(\mathbf{H}^t); \quad \hat{\mathbf{X}}^{t} = \hat{\mathbf{X}}^{t-1} + \Delta \hat{\mathbf{X}}^{t} \) \Comment{Readout}
        \If{\( t \geq 1 \)} \Comment{Prediction phase}
            \State Store \( \hat{\mathbf{X}}^t \); \quad \( \nabla \hat{\mathbf{X}}_{\mathbf{M}}^t = \text{Linear}(\mathbf{M}^t) \) \Comment{STD Readout}
            \State \( \mathcal{L}_{\text{DIC}} \mathrel{+}= \text{DIC}( \nabla \hat{\mathbf{X}}_{\mathbf{M}}^{t-1}, \nabla \hat{\mathbf{X}}_{\mathbf{M}}^t) \) \Comment{DIC}
        \EndIf
    \EndFor
\EndFor

\State \( \mathcal{L} \gets \mathcal{L}_{\text{L1}} + \lambda \mathcal{L}_{\text{DIC}} \); \quad \( \Theta \gets \Theta - \alpha \frac{\partial \mathcal{L}}{\partial \Theta} \) \Comment{Loss update}
\State \Return \( \Theta \) (training) \textbf{or} \( \{\hat{\mathbf{X}}^t\}_{t=1}^T \) (inference)
\end{algorithmic}
\end{algorithm}

\subsection{Core Modules}

\OurModel\ integrates three modules—\textbf{LID}, \textbf{STD}, and \textbf{TAD}—to model local interactions, pollutant transport, and temporal accumulation, ensuring a \textbf{physically consistent and data-driven} framework, as illustrated in Figure~\ref{fig:overall_framework}.

\subsubsection{Local Interaction Dynamics (LID)}

The LID module captures pollutant interactions by embedding \textbf{meteorology ($\mathbf{P}$), emissions ($\mathbf{Q}$), and air pollutant concentrations ($\mathbf{X}$)} into a latent space representation. This process consists of two key stages:

First, the initial hidden state \( \mathbf{H}^t \) is obtained through an embedding step:

\begin{equation}
    \mathbf{H}^t = \text{Linear}([\hat{\mathbf{X}}^{t-1}, \mathbf{P}^t, \mathbf{Q}^t]),
\end{equation}

where \( \hat{\mathbf{X}}^{t-1} \) is the estimated pollutant concentration from the previous step, while \( \mathbf{P}^t \) and \( \mathbf{Q}^t \) represent the meteorological and emissions variables at time \( t \). The Linear layer projects these inputs into the hidden space.

Next, a representation enhancement step refines \( \mathbf{H}^t \) using a multi-layer perceptron (MLP) with normalization:

\begin{equation}
    \mathbf{E}^t = \text{MLP}(\text{RMSNorm}(\mathbf{H}^t)),
\end{equation}

where the MLP applies nonlinear transformations, incorporating \textbf{Linear layers, RMSNorm~\cite{zhang2019root}\footnote{\url{https://pytorch.org/docs/stable/generated/torch.nn.modules.normalization.RMSNorm.html}}, SiLU\footnote{\url{https://pytorch.org/docs/stable/generated/torch.nn.SiLU.html}}, and Dropout} to improve feature expressiveness. This step enhances pollutant interactions, accounting for complex photochemical reactions, emissions-driven variations, and meteorological influences.

\subsubsection{Spatial Transport Dynamics (STD)}

The STD module models \textbf{pollutant transport and dispersion}, capturing advection and diffusion effects across a spatial graph \( \mathcal{G} = (\mathcal{V}, \mathcal{E}) \). Here, each node \( v \in \mathcal{V} \) represents a monitoring station, while edges \( \mathcal{E} \) define spatial connections based on a predefined geodesic distance threshold (e.g., 200 km).

To propagate pollutant states across space, we employ a \textbf{graph convolutional operation}~\cite{kipf2016semi} that aggregates information from neighboring nodes while ensuring numerical stability through a Laplacian-based normalization:

\begin{equation}
    \mathbf{M}^t = \text{Linear}(\tilde{\mathbf{L}} \mathbf{H}^{t}) = \text{Linear}((\mathbf{I} - \mathbf{D}^{-1/2} \mathbf{A} \mathbf{D}^{-1/2}) \mathbf{H}^{t}),
\end{equation}

\noindent where:
\begin{itemize}
    \item \( \mathbf{A} \) is the adjacency matrix, encoding the connectivity between stations.
    \item \( \mathbf{D} \) is the diagonal degree matrix, where \( D_{vv} = \sum_{v' \in \mathcal{N}(v)} A_{vv'} \).
    \item \( \tilde{\mathbf{L}} \) is the \textbf{normalized graph Laplacian}, which ensures balanced information propagation across nodes with varying connectivity.
\end{itemize}

This formulation allows the model to \textbf{capture pollutant dispersion as a second-order process}~\cite{seinfeld2016atmospheric,li2023improving,kipf2016semi}, effectively representing spatial diffusion and advection mechanisms in air quality dynamics. The Laplacian normalization further stabilizes computation across nodes with different degrees, making pollutant transport modeling both scalable and physically meaningful.

To enforce \textbf{mass conservation}~\cite{seinfeld2016atmospheric}, we introduce a \textbf{domain-informed constraint (DIC)} that ensures the net transported pollutant mass remains balanced across all nodes at each time step:

\begin{equation}
    \sum_{v \in \mathcal{V}} \nabla \hat{\mathbf{X}}_{\mathbf{M}}^t = 0,
\end{equation}

where $\nabla \hat{\mathbf{X}}_{\mathbf{M}}^t$ represents the predicted pollutant transport gradient. This loss encourages the model to \textbf{maintain pollutant mass balance}, preventing unphysical accumulation or dissipation.

\subsubsection{Temporal Accumulation Dynamics (TAD)}

The TAD module captures \textbf{long-term pollutant accumulation and decay} by modeling sequential dependencies using a gated recurrent unit (GRU). Given the outputs from LID and STD, the TAD module updates the hidden state as:

\begin{equation}
    \mathbf{Z}^t = \text{GRUcell}(\mathbf{H}^t, \mathbf{Z}^{t-1}).
\end{equation}

This recurrent structure enables \OurModel to track pollution trends over time, accounting for \textbf{pollutant persistence, decay, and meteorology-driven changes}. Unlike transformer-based models that treat time steps as independent tokens, GRU explicitly accumulates past data, improving robustness for long-term forecasting.

\subsection{Prediction Framework and Loss}

As shown in Algorithm~\ref{alg:algorithm}, \OurModel\ employs an iterative prediction strategy to ensure smooth and stable pollutant forecasts. At each time step \( t \), the model first estimates the rate of change in pollutant concentration:

\begin{equation}
    \Delta \hat{\mathbf{X}}^t = \text{Linear}(\mathbf{H}^t).
\end{equation}

The updated pollutant concentration is then obtained using a \textbf{residual formulation}, refining predictions through step-wise accumulation:

\begin{equation}
    \hat{\mathbf{X}}^t = \hat{\mathbf{X}}^{t-1} + \Delta \hat{\mathbf{X}}^t.
\end{equation}

This formulation enables the model to iteratively correct its predictions, preserving physical consistency while mitigating abrupt fluctuations. By modeling pollutant evolution as a sequential adjustment process, \OurModel\ achieves stable and reliable long-term forecasts.

\subsubsection{Loss Functions and Mass Conservation Constraints}

To ensure accurate predictions while maintaining consistency with atmospheric processes, \OurModel\ optimizes the following objective:

\begin{equation}
    \mathcal{L} = \mathcal{L}_{\text{L1}} + \lambda \mathcal{L}_{\text{DIC}}.
\end{equation}

\noindent where:

\begin{itemize}
    \item \(\mathcal{L}_{\text{L1}}\) is the \textbf{prediction loss}, minimizing the absolute error between predicted and observed pollutant concentrations:
    \begin{equation}
        \mathcal{L}_{\text{L1}} = \frac{1}{N T} \sum_{n=1}^N \sum_{t=1}^T \|\hat{\mathbf{X}}_n^t - \mathbf{X}_n^t\|_1.
    \end{equation}
    
    \item \(\mathcal{L}_{\text{DIC}}\) enforces \textbf{mass conservation} constraints on the STD module, ensuring physically consistent transport of pollutants.
\end{itemize}

\subsubsection{Enforcing Mass Conservation in STD}

In atmospheric sciences, pollutant transport follows the \textbf{continuity equation}~\cite{jacobson1999fundamentals}, ensuring pollutants are transported without artificial mass gain or loss. To align with this principle, we explicitly extract the \textbf{spatial transport contribution} from the STD module at each prediction step:

\begin{equation}
    \nabla \hat{\mathbf{X}}_{\mathbf{M}}^t = \text{Linear}(\mathbf{M}_t).
\end{equation}

This term explicitly represents the \textbf{modeled pollutant transport effect} from the STD module. To ensure mass balance across both space and time, \OurModel\ applies a \textbf{domain-informed constraint (DIC)} that enforces continuity in pollutant transport:

\begin{equation}
    \mathcal{L}_{\text{DIC}} = \sum_{t=1}^{T} \|\nabla \hat{\mathbf{X}}_{\mathbf{M}}^t - \nabla \hat{\mathbf{X}}_{\mathbf{M}}^{t-1}\|_2.
\end{equation}

To further ensure that pollutant transport strictly follows physical principles, we impose two \textbf{mass conservation constraints}:

\begin{itemize}
    \item \textbf{Spatial Mass Conservation} At each step, the total pollutant mass exchanged between connected nodes in the station graph (\(\mathcal{G}\)) must sum to zero, preserving balance and preventing artificial generation or dissipation of pollutants:
    \begin{equation}
        \sum_{v' \in \mathcal{N}(v)} \nabla \hat{\mathbf{X}}_{\mathbf{M}, v \to v'}^t = 0, \quad \forall v \in \mathcal{V}.
    \end{equation}
    Here, \( \nabla \hat{\mathbf{X}}_{\mathbf{M}, v \to v'}^t \) represents the pollutant flux from node \( v \) to its neighbor \( v' \), directly extracted from STD. This ensures that pollutants transported away from a node must reappear in its neighboring nodes.

    \item \textbf{Temporal Mass Conservation} The total pollutant mass across all nodes should remain stable over time, preventing artificial sources or sinks:
    \begin{equation}
        \frac{d}{dt} \sum_{v \in \mathcal{V}} \nabla \hat{\mathbf{X}}_{\mathbf{M}, v}^t = 0.
    \end{equation}
\end{itemize}

\subsubsection{DIC Loss Formulation}

To integrate these constraints into training, we define the final domain-informed loss:

\begin{equation}
\mathcal{L}_{\text{DIC}} = \frac{1}{|\mathcal{V}|} \sum_v \left| \sum_{v’ \in \mathcal{N}(v)} \nabla \hat{\mathbf{X}}_{\mathbf{M}, v \to v’}^t \right|
+ \frac{1}{T} \sum_t \left| \sum_v \frac{d \nabla \hat{\mathbf{X}}_{\mathbf{M}, v}^t}{dt} \right|.
\end{equation}

\subsubsection{Interplay Between STD and GRU in \OurModel}

While the STD module is explicitly constrained to enforce pollutant transport physics, \textbf{pollutant accumulation and secondary transformations} (e.g., chemical reactions, deposition effects) are handled by the GRU-based TAD. The GRU is responsible for learning time-dependent variations, including pollutant persistence, atmospheric reactions, and slow temporal drifts, while the STD ensures instantaneous pollutant transport follows physical constraints.

By constraining pollutant movement within STD and allowing GRU to model long-term temporal dynamics, \OurModel\ achieves \textbf{physically grounded yet flexible pollutant forecasting}, ensuring improved generalization and interpretability.

\subsection{Deployment Framework}

The deployment pipeline for \OurModel\ ensures real-time readiness, scalability, high availability, and concurrency, as illustrated in Figure~\ref{fig:overall_framework}. Preprocessed inputs, including historical air quality, meteorological forecasts (e.g., GFS), and emissions, are fed into \OurModel, which outputs pollutant predictions via a Flask and NginX API framework. Containerized with Docker and orchestrated through Kubernetes, the system supports dynamic scaling across cloud platforms like AWS and Aliyun. To ensure high availability and concurrency, redundant deployments, load balancing, and asynchronous endpoints are employed, while meteorological updates, provided four times daily, align predictions with real-time needs.

\section{Experiments}
\label{sec:experiments}

This section evaluates the effectiveness of \OurModel\ for air quality forecasting (AQF) using real-world datasets. We detail the experimental setup, including dataset configuration, implementation details, and baseline comparisons, followed by a comprehensive analysis of results.

\subsection{Experimental Setup}

\subsubsection{Dataset}
To ensure robust evaluation, we construct a large-scale dataset spanning 2016–2023, incorporating air quality, meteorological, and emissions data. The dataset includes a total of \textbf{70,128 hours} of observations from \textbf{355 monitoring stations}. It comprises \textbf{2 air quality variables} (e.g., \PM and \ozone), \textbf{6 emissions variables} (e.g., \NitrogenOxides, \VOC), and \textbf{8 meteorological variables} (e.g., temperature, wind components). The years 2016–2019 are used for training, 2020–2021 for validation and hyperparameter tuning, and 2022-2023 for testing. This split ensures fair evaluation of model generalization. A detailed summary of the dataset variables is provided in Appendix~\ref{appendix:variables}.

\subsubsection{Implementation Details}

\OurModel\ is implemented using PyTorch, with graph-based computations handled by PyTorch Geometric (PyG)~\cite{fey2019fast}. Training is conducted on an Nvidia 4070S GPU, enabling efficient large-scale experiments. The model is optimized for deployment on CPUs, ensuring computational efficiency and cost-effectiveness.

The training process employs the Adam optimizer with an initial learning rate of $1 \times 10^{-4}$, a batch size of 32, and runs for 100 epochs. Early stopping is applied based on validation MAE to prevent overfitting, and a learning rate scheduler (\texttt{ReduceLROnPlateau}) is used for dynamic adjustment. To improve computational efficiency, mixed precision training with FP16 is enabled.

\paragraph{Evaluation Metrics}
To quantitatively assess the forecasting accuracy of \OurModel, we employ \textbf{Mean Absolute Error (MAE)} and \textbf{Root Mean Square Error (RMSE)}. MAE provides an interpretable measure of absolute prediction error, while RMSE penalizes larger deviations, making it more sensitive to extreme prediction errors. The definitions of these metrics are provided in Appendix~\ref{appendix:metrics}.

\subsubsection{Baselines}

We evaluate \OurModel\ against three categories of baseline models, with Table~\ref{tab:methods} in Appendix~\ref{appendix:models} summarizing their key modeling characteristics. Additionally, Figure~\ref{fig:input_output} in Appendix~\ref{appendix:frameworks} illustrates their input-output frameworks based on their original designs.

\begin{itemize}
    \item \textbf{Machine Learning Models}: These include \textbf{XGBoost}~\cite{chen2016xgboost} and \textbf{LightGBM}~\cite{ke2017lightgbm}, which directly fit air quality predictions to meteorological and emissions data using gradient boosting but lack temporal and spatial modeling.
    
    \item \textbf{General Time-Series Forecasting Models}: These include transformer-based architectures such as \textbf{iTransformer}~\cite{liu2023itransformer}\footnote{\url{https://github.com/thuml/iTransformer}} and \textbf{TimeXer}~\cite{wang2024timexer}\footnote{\url{https://github.com/thuml/TimeXer}}, which model historical air quality, meteorology, and emissions in the encoder but predict air quality directly in the decoder without incorporating forecasted meteorological data.
    
    \item \textbf{Air Quality Forecasting (AQF) Models}: These methods are specifically tailored for air pollution forecasting, integrating spatiotemporal and, in some cases, physical constraints. Notable examples include \textbf{GC-LSTM}~\cite{qi2019hybrid}, a hybrid model combining graph convolution (GCN) and LSTM for spatiotemporal prediction; \textbf{\PM-GNN}~\cite{wang2020pm25}\footnote{\url{https://github.com/shuowang-ai/PM2.5-GNN}}, a GNN-based model incorporating meteorological inputs but lacking explicit physical constraints; and \textbf{AirPhyNet}~\cite{hettige2024airphynet}\footnote{\url{https://github.com/kethmih/AirPhyNet}}, a physics-informed neural ODE~\cite{chen2018neural} model using historical air quality and wind to construct advection-diffusion kernels, without other meteorological or emissions inputs.
    
\end{itemize}

\begin{table*}[t!]
\centering
\caption{Performance comparison of models on BTHSA and YRD data in a 72-hour forecasting task. The reported metrics are the average RMSE and MAE across all time steps compared with the ground truth.}
\label{tab:performance}
\begin{tabular}{l|cc|cc|cc|cc}
\toprule
\multirow{2}{*}{Model (Publication)} & \multicolumn{4}{c|}{BTHSA} & \multicolumn{4}{c}{YRD} \\ 
\cmidrule(lr){2-5} \cmidrule(lr){6-9}
& \multicolumn{2}{c|}{\textbf{RMSE} $\downarrow$} & \multicolumn{2}{c|}{\textbf{MAE} $\downarrow$}  
& \multicolumn{2}{c|}{\textbf{RMSE} $\downarrow$} & \multicolumn{2}{c}{\textbf{MAE} $\downarrow$} \\  
& \PM & \ozone & \PM  & \ozone  & \PM & \ozone & \PM & \ozone \\ 
\midrule
XGBoost \cite{chen2016xgboost} (NeurIPS'16) & 35.47 & \underline{27.53} & 26.05 & \underline{20.68} & 22.07 & \underline{27.71} & 16.64 & \underline{20.99} \\ 
LightGBM \cite{ke2017lightgbm} (NeurIPS'17) & 35.48 & 27.97 & 26.07 & 21.05 & 22.06 & 28.07 & 16.74 & 21.33 \\ 
GC-LSTM \cite{qi2019hybrid} (STOTEN'19) & 32.95 & 35.12 & 23.09 & 26.82 & 18.77 & 34.65 & 13.57 & 26.31 \\ 
\PM-GNN \cite{wang2020pm25} (SIGSPATIAL'20) & 31.51 & 31.85 & 23.5 & 22.9 & 18.17 & 31.46 & 13.24 & 24.05 \\ 
iTransformer \cite{liu2023itransformer} (ICLR'23) & \underline{31.17} & 29.9 & 20.72 & 22.58 & 17.92 & 30.53 & \underline{12.86} & 23.28 \\ 
AirPhyNet \cite{hettige2024airphynet} (ICLR'24) & 50.49 & 68.04 & 40.26 & 51.53 & 28.53 & 56.96 & 22.9 & 42.76 \\ 
TimeXer \cite{wang2024timexer} (NeurIPS'24) & 31.51 & 30.07 & \underline{20.6} & 22.87 & \underline{17.89} & 30.47 & \underline{12.86} & 23.28 \\ 
\midrule
\textbf{\OurModel(Ours)} & \textbf{24.13} & \textbf{22.45} & \textbf{15.46} & \textbf{16.73} 
& \textbf{15.55} & \textbf{23.03} & \textbf{10.97} & \textbf{17.27} \\ 
\bottomrule
\end{tabular}
\end{table*}

\begin{figure}[t]
  \centering
  \includegraphics[width=\linewidth]{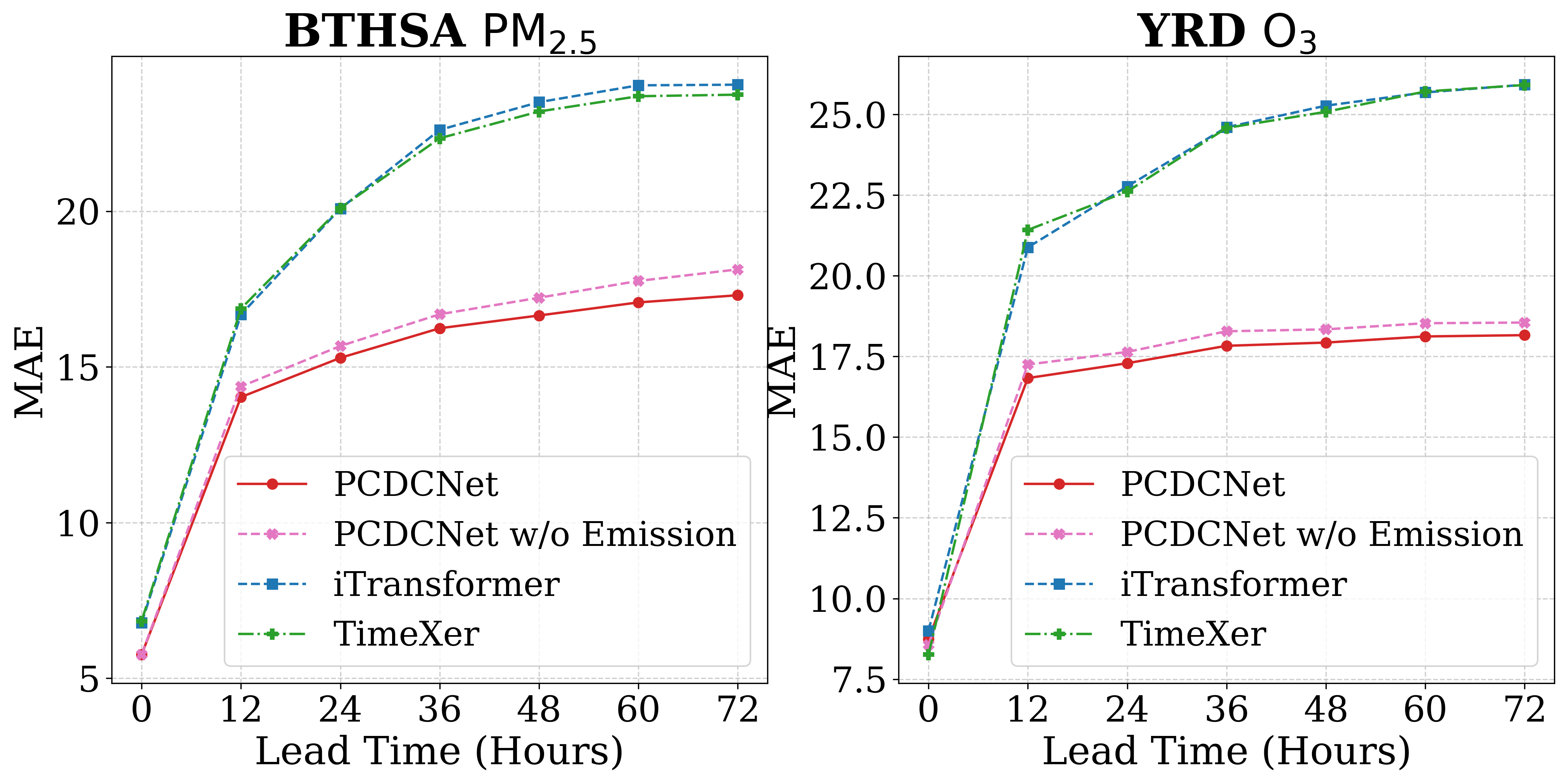}
    \caption{MAE trends for \(\mathrm{PM}_{2.5}\) in BTHSA and \(\mathrm{O}_3\) in YRD over a 72-hour prediction horizon.}
  \Description{}
  \label{fig:leadtime}
\end{figure}

\subsection{Performance Analysis}

Table~\ref{tab:performance} compares \OurModel\ with both traditional (e.g., XGBoost, LightGBM) and advanced (e.g., iTransformer, TimeXer) models on a 72-hour AQF task in the BTHSA and YRD regions. \OurModel consistently achieves the lowest MAE across both pollutants and regions. Notably, it reduces the prediction error by \textbf{19.8\%} for \PM and \textbf{18.4\%} for \ozone compared to the next best method. This improvement is especially pronounced for long lead times of 72, underscoring the model’s robustness in scenarios where baseline methods often experience accumulating errors.

Figure~\ref{fig:leadtime} further illustrates these gains by showing the MAE trends for \PM in BTHSA and \ozone in YRD over the full 72-hour horizon. While competing approaches see their errors increase sharply with extended lead times, \OurModel\ maintains relatively stable performance, demonstrating its capacity to retain predictive accuracy even under longer forecasting windows.

\paragraph{CMAQ Input-Output Framework}
These results strongly echo our argument from the Introduction that a surrogate model grounded in numerical modeling principles can offer both computational efficiency and physical consistency. By aligning the input-output structure of \OurModel\ with how CMAQ (and similar systems) incorporate meteorological forecasts and emissions data, we capture the multi-scale dynamics underpinning air quality. This design choice yields a meaningful reduction in errors over purely time-series-based architectures, particularly for secondary pollutants like \ozone, which are sensitive to meteorological variations and photochemical reactions.

\paragraph{Error Reductions Through Emissions}
Unlike general-purpose transformer models, \OurModel\ explicitly integrates forecasted meteorology and emissions. This focus on domain-specific drivers yields error reductions of \textbf{9.8\%} in \PM and \textbf{3.7\%} in \ozone predictions over baselines lacking emissions data (Figure~\ref{fig:emissions_ablation} in Appendix~\ref{appendix:emissions}). These improvements validate the need for pollutant-specific modeling strategies where meteorological conditions and emissions intricacies are paramount.

\paragraph{Long-Term Stability and Physical Consistency}
Finally, the stable performance at extended lead times aligns with the emphasis on bridging physics-based insights with data-driven flexibility. By maintaining a structured framework that enforces mass conservation and accounts for transport processes, \OurModel\ mitigates error propagation and delivers physically plausible forecasts over 72 hours. In doing so, it serves as a computationally efficient surrogate that remains interpretable and reliable—a critical requirement for real-time air quality management and decision-making.

\begin{figure*}[h]
  \centering
  \includegraphics[width=\linewidth]{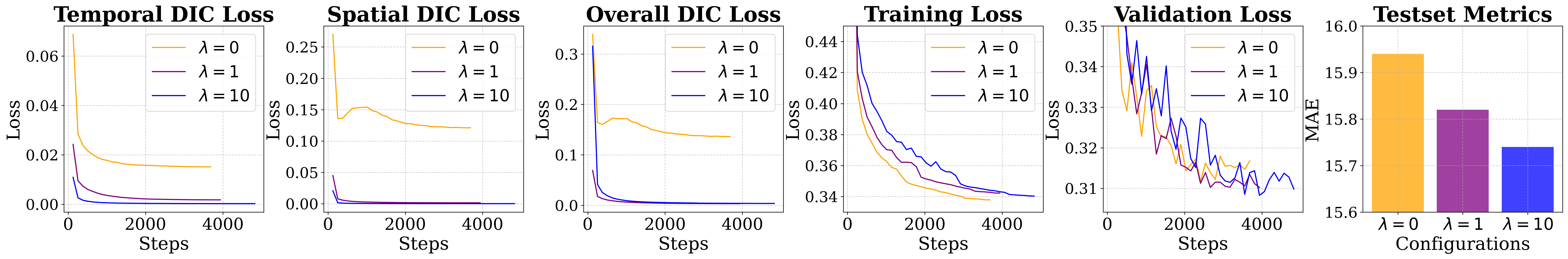}
  \caption{Analysis of Domain-Informed Constraints (DIC). Temporal and spatial DIC loss components confirm the necessity of separately modeling pollutant conservation across time and space. Stronger DIC constraints (\(\lambda = 10\)) improve test set performance while maintaining physical consistency.}
  \Description{}
  \label{fig:dic_analysis}
\end{figure*}

\subsection{Ablation Studies and Parameter Sensitivity}

\begin{figure}[h]
  \centering
  \includegraphics[width=\linewidth]{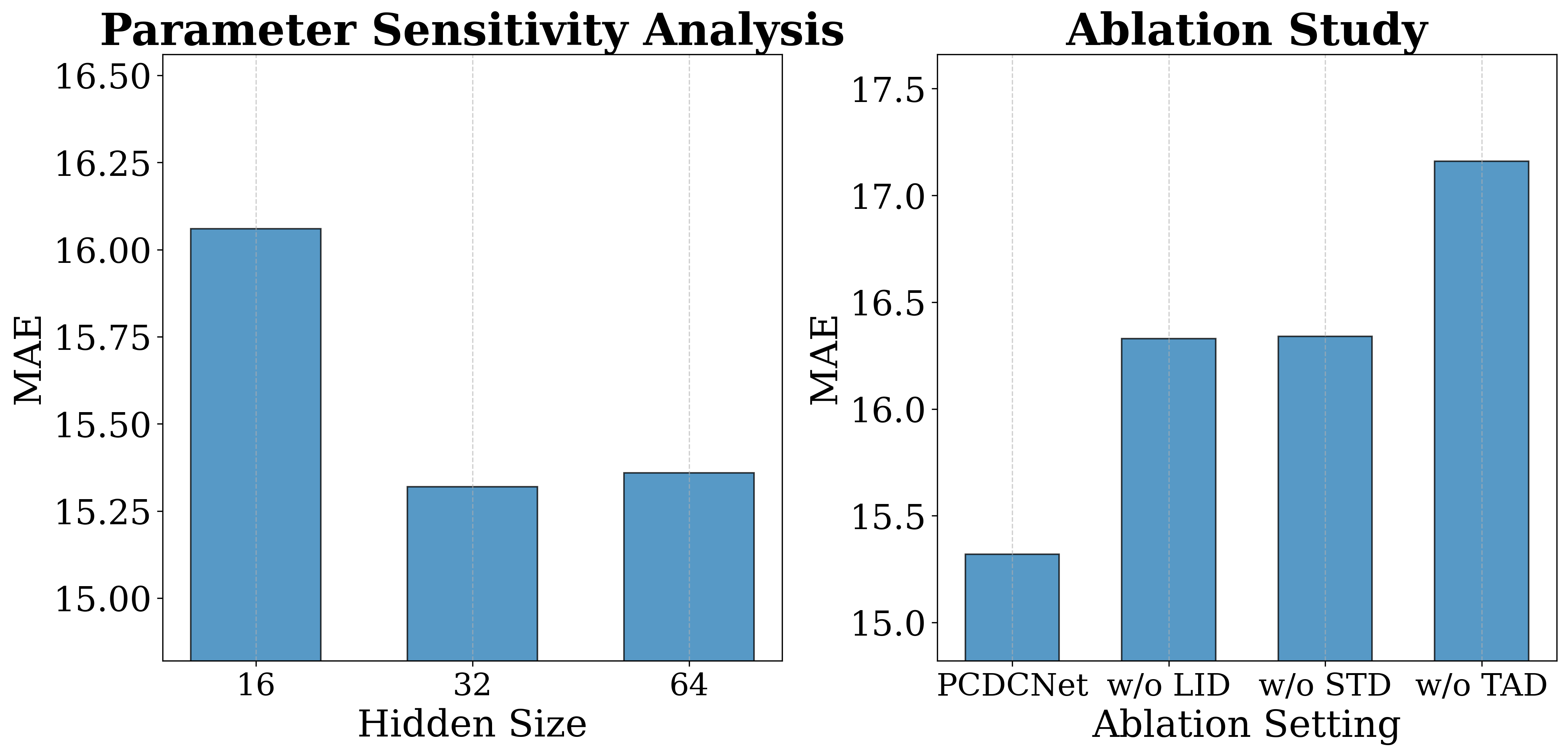}
    \caption{\textbf{Performance evaluation in BTHSA.} (Left) Sensitivity analysis of hidden size (\textbf{16, 32, 64}) shows that \textbf{32} yields the lowest MAE, balancing complexity and generalization. (Right) Ablation study confirms performance drops when removing components, validating the model design.}
  \Description{}
  \label{fig:ablation_sensitivity}
\end{figure}

To evaluate the contributions of individual components and hyperparameter choices in \OurModel, we conduct ablation studies and sensitivity analysis, as shown in Figure~\ref{fig:ablation_sensitivity}.

\paragraph{Ablation Studies}
Ablation results demonstrate the importance of \textbf{LID}, \textbf{STD}, and \textbf{TAD}. Removing \textbf{LID} weakens the model's ability to capture nonlinear interactions, while excluding \textbf{STD} impacts pollutant transport modeling. The largest performance drop occurs when \textbf{TAD} is omitted, underscoring its critical role in capturing temporal dependencies and propagating historical information.

\paragraph{Parameter Sensitivity}
Sensitivity analysis shows that a hidden size of 32 achieves the best balance between model capacity and computational efficiency, outperforming smaller sizes prone to underfitting and larger sizes prone to overfitting. These results validate \OurModel's design and its ability to achieve robust air quality forecasting accuracy.

\subsection{Physical Consistency and Generalization}

Figure~\ref{fig:dic_analysis} illustrates the impact of the Domain-Informed Constraints (DIC) loss under varying strengths of the hyperparameter \(\lambda\), where \(\lambda = 0\) indicates the absence of DIC loss (only \(\mathcal{L}_{\text{L1}}\) is used), and \(\lambda = 10\) applies stronger constraints.

\paragraph{Impact on Test Set Performance}
As \(\lambda\) increases, the test set MAE consistently decreases, demonstrating the effectiveness of the DIC loss in improving generalization. For \(\lambda = 0\), the model lacks explicit constraints to enforce physical consistency, leading to suboptimal performance. Conversely, \(\lambda = 10\) achieves the best results by strongly guiding the model toward domain-consistent behavior.

\paragraph{Implicit Mass Conservation in STD}
Interestingly, even when \(\lambda = 0\), the DIC loss curves (temporal, spatial, and overall) exhibit a rapid initial decrease (Figure~\ref{fig:dic_analysis}). This suggests that the STD module inherently learns to approximate mass conservation, despite the absence of constraints. However, without DIC loss, values stabilize at higher levels, indicating insufficient guidance for near-zero physical inconsistencies. When \(\lambda > 0\), the DIC loss reduces significantly, validating the integration of domain knowledge into training.

\paragraph{Training and Validation Loss Dynamics}
The training loss is lowest for \(\lambda = 0\), as the absence of constraints allows the model to overfit. However, the validation loss shows the opposite trend: stronger DIC constraints (\(\lambda = 10\)) yield better generalization. This inversion underscores the role of DIC loss in improving robustness and alignment with physical laws.

\paragraph{Component-Wise Effectiveness}
The temporal and spatial components of the DIC loss validate the necessity of separately modeling these two aspects of pollutant dynamics. Temporal DIC ensures consistency across time steps, while spatial DIC enforces pollutant conservation across neighboring nodes, together forming a comprehensive representation of atmospheric transport dynamics.

These results validate the design and integration of DIC loss into \OurModel, showcasing its effectiveness in embedding domain knowledge and enhancing predictive accuracy and interpretability.

\subsection{Online Deployment and Case Studies}

\paragraph{Real-Time Deployment}
\OurModel has been deployed as a real-time AQF system\footnote{\url{https://caiyunapp.com/map/}}. Forecasted meteorological inputs, updated every six hours, are sourced from \textbf{GFS}\footnote{\url{https://www.ncei.noaa.gov/products/weather-climate-models/global-forecast}} or \textbf{EC}\footnote{\url{https://www.ecmwf.int/}}.

\paragraph{Case Studies}

\OurModel was evaluated on major pollution events, including Spring Festival haze, severe Beijing pollution, and California wildfires, consistently providing early warnings and capturing pollution dynamics (Appendix~\ref{appendix:case_studies}).

\section{Conclusion}

This work introduces \OurModel, the Physical-Chemical Dynamics and Constraints Network, a scalable and interpretable surrogate model bridging deep learning efficiency with the physical consistency of numerical systems like CMAQ. By integrating emissions data, historical observations, and forecasted meteorology, \OurModel captures complex spatiotemporal and nonlinear interactions through its LID, STD, and TAD modules, while Domain-Informed Constraints enforce alignment with atmospheric principles. Extensive experiments validate its state-of-the-art performance in \PM and \ozone forecasting, demonstrating robust generalization and physical consistency. Future work will focus on optimizing scalability, refining chemical mechanisms, and expanding its application to larger or global regions, advancing air quality forecasting.

\begin{acks}
This work was supported by the National Natural Science Foundation of China (Grant No. 42450183, 12275020, 12135003, 12205025, 42461144209), the Ministry of Science and Technology of China (2023YFE0109000). Jingfang Fan is supported by the Fundamental Research Funds for the Central Universities. Shuo Wang acknowledges the financial support by China Scholarship Council (CSC) Grant No. 202106040117.
\end{acks}

\clearpage
\bibliographystyle{ACM-Reference-Format}
\bibliography{PCDCNet-reference}

\clearpage
\appendix

\section{Appendices}

\subsection{Related Work}

\subsubsection{Numerical Models for AQF}
Numerical models like CMAQ~\cite{appel2017description}, WRF-Chem~\cite{sicard2021high}, and CAMx~\cite{emery2024comprehensive} are widely used for AQF. They solve PDEs to simulate pollutant dynamics. Their interpretability and alignment with physical and chemical principles make them invaluable for understanding pollution transport and formation mechanisms. However, these models heavily rely on emission inventories, which are often uncertain, and their high computational cost renders them impractical for real-time applications. Furthermore, these models are not inherently designed to assimilate real-time observational data, limiting their ability to capture dynamic events like sudden pollution spikes~\cite{park2021implementation}, and reducing their effectiveness for timely forecasting. Despite these limitations, their robustness and interpretability offer valuable guidance for integrating domain knowledge into AI models.

\subsubsection{Deep Learning Models for AQF}
Deep learning has become a promising alternative, offering computational efficiency and the ability to model complex patterns in observational data. Methods such as Pangu-Weather~\cite{bi2023accurate} and GraphCast~\cite{lam2023learning} leverage AI for meteorological forecasting but are not designed for AQF, often ignoring emissions and pollutant-specific dynamics. Within AQF, recurrent models (e.g., GRU, LSTM) capture temporal patterns, while graph-based methods (e.g., GC-LSTM~\cite{qi2019hybrid}, \PM-GNN~\cite{wang2020pm25}) model spatial dependencies. 
AirPhyNet~\cite{hettige2024airphynet} incorporates physics-informed modeling using neural ordinary differential equations (neural ODEs)~\cite{chen2018neural} to simulate pollutant dynamics like advection and diffusion. While this approach enhances interpretability and aligns predictions with physical principles, it is limited by the inherent constraints of neural ODEs, such as the inability to integrate additional meteorological variables effectively. This limitation reduces its flexibility and accuracy in capturing the multifaceted interactions of emissions, meteorology, and pollutants.

These models face challenges in handling varying pollution contexts. For instance, models trained on high-pollution regions often underperform in low-pollution scenarios due to their reliance on data-driven learning. Moreover, many deep learning approaches overlook emissions data or fail to incorporate domain knowledge, limiting their interpretability and accuracy in specialized AQF tasks.

\subsubsection{Mixing Numerical Models \& Deep Learning}
The integration of numerical and deep learning models aims to combine their respective strengths. Numerical models provide interpretability and adherence to atmospheric principles but are computationally expensive. Deep learning methods, while efficient and capable of fitting observational data, often lack generalizability and domain-specific insights. Hybrid approaches, such as Aurora~\cite{bodnar2024aurora}, attempt to bridge this gap but are limited by their dependence on structured reanalysis data, lack of emissions modeling, and absence of station-level observations.

\OurModel addresses these challenges by combining emissions, meteorology, and physical-chemical constraints in a unified framework. It integrates the interpretability of numerical models with the efficiency of deep learning, enabling accurate and scalable AQF. By explicitly modeling pollutant interactions and leveraging domain-informed constraints, \OurModel provides a robust solution for real-time, station-level air quality forecasting.

\subsection{Research Regions and Spatial Graph Construction}

\begin{figure}[h]
  \centering
  \includegraphics[width=0.9\linewidth]{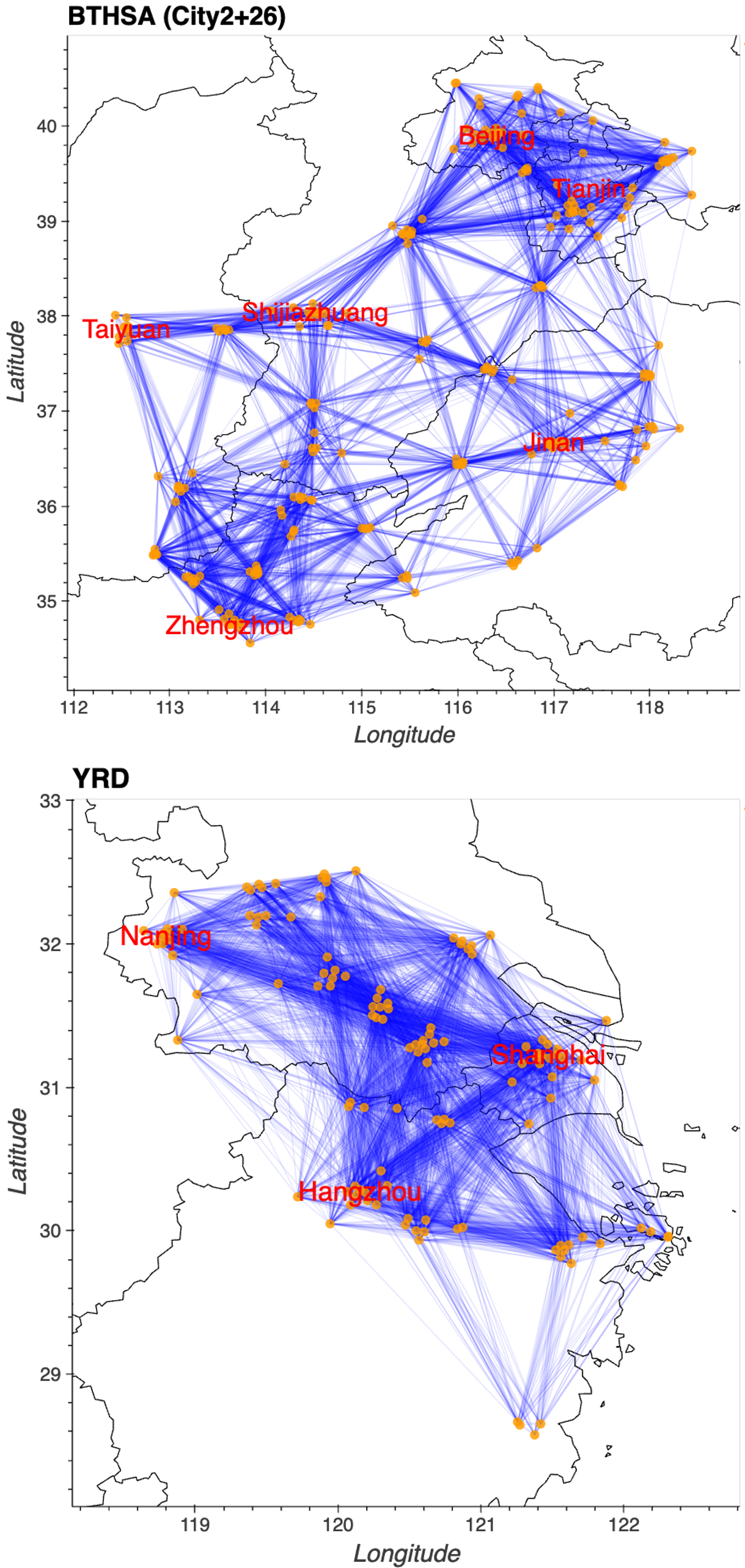}
  \caption{Research regions and constructed spatial graphs for the Beijing-Tianjin-Hebei and Surrounding Areas (BTHSA, top) and the Yangtze River Delta (YRD, bottom). Each orange node represents a monitoring station, and blue edges indicate spatial connections based on a 200 km geodesic distance threshold. Major cities are labeled in red, highlighting their central role in the regional graph structure. These graphs capture the spatial dependencies necessary for pollutant transport modeling and serve as the basis for graph-based components in our framework.}
  \Description{}
  \label{fig:regions_graph}
\end{figure}

To ensure the effectiveness of spatial modeling in \OurModel, we focus on two key regions in China: the Beijing-Tianjin-Hebei and Surrounding Areas (BTHSA) and the Yangtze River Delta (YRD). These regions are selected due to their distinct meteorological patterns, pollutant transport dynamics, and high-density air quality monitoring networks.

Figure~\ref{fig:regions_graph} illustrates the constructed spatial graphs for these regions. Each orange node represents an air quality monitoring station, while blue edges denote spatial connections between nodes based on a geodesic distance threshold of 200 km. Major cities are highlighted in red, emphasizing their significant roles in regional pollutant transport and emissions.

The spatial graph captures critical dependencies for pollutant transport, including horizontal advection and diffusion, by linking nearby monitoring stations. These connections form the foundation for the graph-based components in \OurModel, enabling effective spatiotemporal modeling. This graph-based structure ensures that both localized and regional pollutant dynamics are captured, aligning with real-world atmospheric processes.

\subsection{Overview of Variables}
\label{appendix:variables}
The variables used in this study are categorized into three primary groups: \textbf{Air Pollutants}, \textbf{Meteorological Factors}, and \textbf{Emissions Data}. These variables capture key aspects of air quality dynamics, including pollutant concentrations, atmospheric conditions, and anthropogenic emissions, ensuring a robust input space for modeling.

\subsubsection{Air Pollutants (\AirPoll)}
Air pollutant variables represent the primary targets of the forecasting model, capturing concentrations of key pollutants:
\begin{itemize}
    \item \textbf{\(\mathrm{PM}_{2.5}\)} Fine particulate matter with a diameter of 2.5 micrometers or smaller (\(\mu g/m^3\)).
    \item \textbf{\(\mathrm{O}_3\)} Ground-level \ozone (\(\mu g/m^3\)), a secondary pollutant influenced by precursor emissions and meteorological conditions.
\end{itemize}

\subsubsection{Meteorological Factors (\Meteo)}
Meteorological variables describe the atmospheric conditions influencing pollutant dispersion, chemical reactions, and deposition:
\begin{itemize}
    \item \textbf{t2m} 2-meter air temperature (\(K\)).
    \item \textbf{d2m} 2-meter dew point temperature (\(K\)).
    \item \textbf{tp} Total precipitation (\(m\)).
    \item \textbf{sp} Surface pressure (\(Pa\)).
    \item \textbf{blh} Boundary layer height (\(m\)), a key factor in vertical mixing.
    \item \textbf{msdwswrf (swr)} Mean surface downward shortwave radiation flux (\(W/m^2\)), influencing photochemical reactions like \ozone formation.
    \item \textbf{u100, v100} Wind components at 100m altitude (\(m/s\)), critical for pollutant advection.
\end{itemize}

\subsubsection{Emissions Data (\Emiss)}
Emissions data quantify anthropogenic contributions to air quality and include:
\begin{itemize}
    \item \textbf{\(\mathrm{PM}_{2.5}\)} Fine particulate matter emissions (\(ton\)).
    \item \textbf{\(\mathrm{PM}_{10}\)} Coarse particulate matter emissions (\(ton\)).
    \item \textbf{\(\mathrm{NO}_x\)} Nitrogen oxides (\(ton\)), precursors to \ozone and nitrate aerosol.
    \item \textbf{\(\mathrm{VOC}\)} Volatile organic compounds (\(ton\)), precursors to \ozone.
    \item \textbf{\(\mathrm{NH}_3\)} Ammonia (\(ton\)), contributing to secondary particulate formation.
    \item \textbf{\(\mathrm{SO}_2\)} Sulfur dioxide (\(ton\)), a precursor to sulfate aerosols.
\end{itemize}

\subsubsection{Data Sources}
The data were compiled from multiple authoritative sources, including:
\begin{itemize}
    \item \textbf{Meteorology} ERA5 reanalysis data for high-resolution atmospheric parameters.
    \item \textbf{Emissions} MEIC (Multi-resolution Emission Inventory for China) for emissions statistics.
    \item \textbf{Air Pollutants} Observations from ground-based monitoring stations across China.
\end{itemize}

\subsection{Notation Table}
To aid in understanding the methodology and components of \OurModel, Table~\ref{tab:symbols} summarizes the key symbols and their corresponding descriptions used throughout the paper.

\begin{table}[h!]
\centering
\caption{Summary of Symbols Used in the Methodology}
\label{tab:symbols}
\begin{tabular}{@{}p{0.15\linewidth}p{0.75\linewidth}@{}}
\toprule
\textbf{Symbol} & \textbf{Description} \\       
\midrule

$\mathbf{X}^t, \hat{\mathbf{X}}^t$ & Observed and predicted pollutant concentrations (e.g., \PM, \ozone) at time \(t\). \\
$\mathbf{P}^t, \mathbf{Q}^t$ & Meteorological and emission variables (e.g., $\mathbf{P}^t_{\mathrm{t2m}}$ for temperature, $\mathbf{Q}^t_{\mathrm{NO}_x}$ for NO\(_x\) emissions). \\
$T', T$ & Historical window (\(T'\)) and prediction window (\(T\)) lengths. \\
$\mathcal{G} = (\mathcal{V}, \mathcal{E})$ & Station graph: $\mathcal{V}$ (nodes), $\mathcal{E}$ (edges). \\
$v, v', \mathcal{N}(v)$ & A station node \(v\), its neighboring node \(v'\), and the set of all neighbors \(\mathcal{N}(v)\). \\
\( \mathcal{F}_\Theta \) & The model with learnable parameters \(\Theta\), integrating meteorology, emissions, and spatiotemporal dynamics. \\
$\mathbf{H}^t, \mathbf{E}^t$ & Hidden states where \(\mathbf{H}^t\) represents the rate of change in pollutant, meteorology, and emissions interactions, while \(\mathbf{E}^t\) captures localized physical-chemical interactions. \\
$\mathbf{M}^t, \hat{\mathbf{X}}_{\mathbf{M}}^t$ & Aggregated message (\(\mathbf{M}^t\)) for pollutant transport and its readout (\(\hat{\mathbf{X}}_{\mathbf{M}}^t\)) via STD module. \\
$ \mathcal{L}_{\text{DIC}} $ & Domain-Informed Constraint loss for enforcing physical consistency. \\
$\mathbf{u}, \nabla, \Delta$ & Wind vector (\(\mathbf{u}\)), spatial gradient operator (\(\nabla\)) for pollutant advection and diffusion, and temporal difference operator (\(\Delta\)) representing changes along the time dimension. \\
$d_{vv'}$ & Geodesic distance between stations \(v\) and \(v'\). \\
\bottomrule
\end{tabular}
\end{table}

\subsection{Input-Output Frameworks in Air Quality Prediction}
\label{appendix:frameworks}

\begin{figure}[h]
  \centering
  \includegraphics[width=\linewidth]{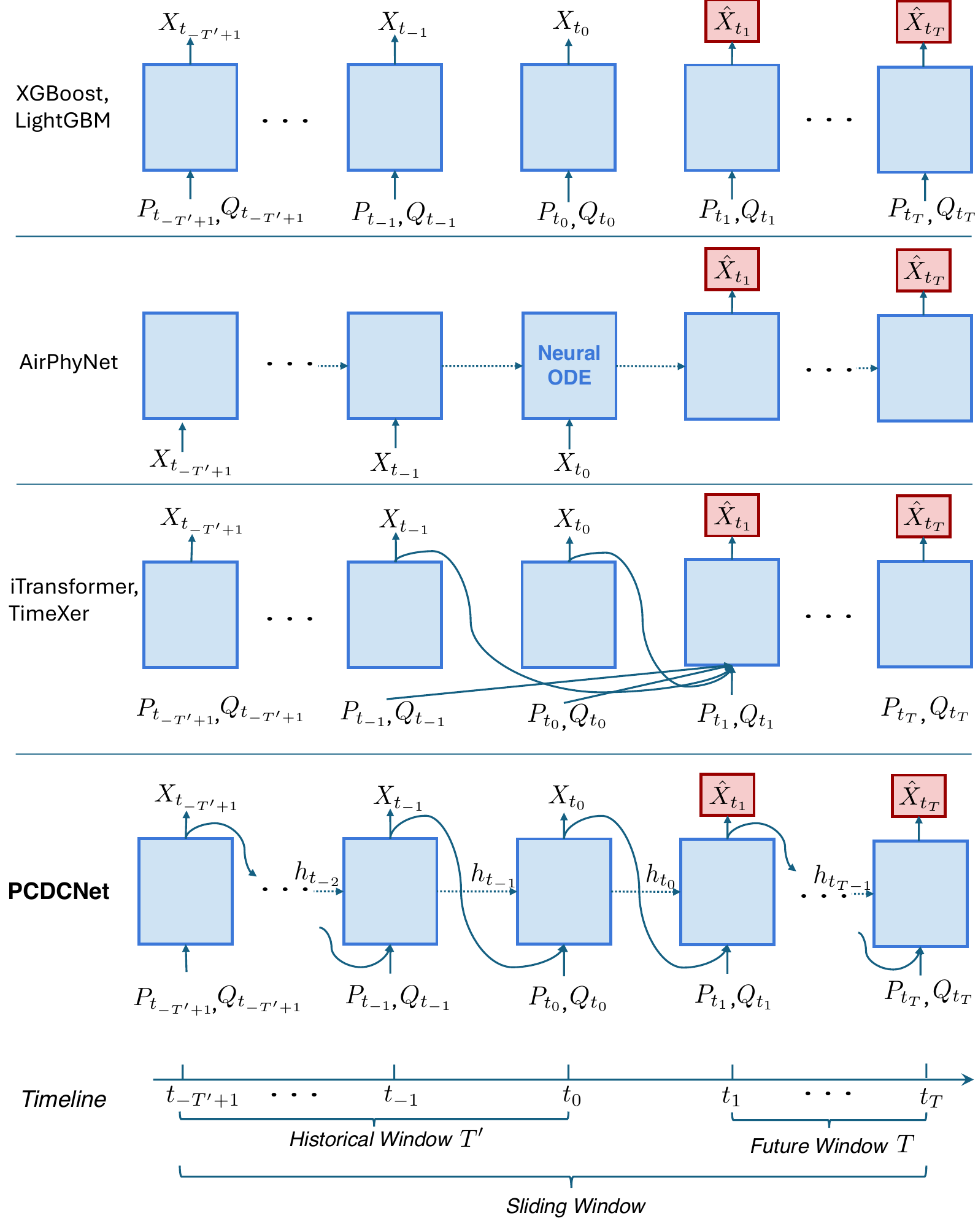}
  \caption{Comparison of input-output frameworks in air quality prediction. XGBoost and LightGBM process each time step independently. AirPhyNet links inputs sequentially via NeuralODEs \cite{chen2018neural}. Transformer-based models (e.g., iTransformer, TimeXer) capture dependencies across historical windows. \OurModel integrates emissions, meteorology, and historical air quality using recurrent mechanisms and sliding windows, enabling spatiotemporal and physical-chemical dynamic modeling.}  
  \label{fig:input_output}
\end{figure}

Air quality forecasting (AQF) systems adopt diverse input-output frameworks tailored to their design goals. Figure~\ref{fig:input_output} summarizes the paradigms of representative models:

\begin{itemize}
    \item \textbf{XGBoost and LightGBM}: These models predict pollutant concentrations for each time step independently, leveraging historical meteorology and emissions but lacking temporal dependency modeling.
    \item \textbf{AirPhyNet}: Introduces sequential modeling using NeuralODEs \cite{chen2018neural}, focusing on pollutant dynamics but limited by its inability to incorporate forecasted meteorology or emissions.
    \item \textbf{Transformer-based Models} (e.g., iTransformer, TimeXer): Utilize self-attention for capturing long-range temporal dependencies across historical windows. However, these models often exclude emissions and domain-specific constraints, limiting their interpretability.
    \item \textbf{\OurModel}: A unified framework that incorporates historical and forecasted meteorology, emissions, and pollutant data via recurrent mechanisms and sliding windows. It explicitly integrates spatiotemporal dependencies and physical-chemical dynamics, ensuring accurate predictions.
\end{itemize}

This unified approach aligns with real-world AQF requirements, bridging the gap between computational efficiency and domain-specific modeling, a feature traditional numerical models lack.

\subsection{Baseline Model Comparison}
\label{appendix:models}

Table~\ref{tab:methods} presents a comparison of baseline models based on their core capabilities. The table highlights key features such as whether the model is specifically designed for air quality forecasting (AQF), supports temporal dependencies (Temp), handles multivariate predictions (MultiV), incorporates future exogenous variables (Exog), models spatial correlations (Spat), and integrates physical constraints (Phy). 

Traditional methods like XGBoost and LightGBM lack temporal and spatial modeling capabilities, limiting their suitability for AQF tasks. Advanced models such as AirPhyNet and \PM-GNN introduce spatial and temporal dynamics but fail to fully integrate exogenous variables or enforce physical constraints. On the other hand, \OurModel and CMAQ comprehensively address all listed capabilities, providing robust solutions for accurate and interpretable air quality forecasting.

\begin{table}[h!]
\centering
\caption{
Comparison of baseline models based on their native capabilities. Columns represent: 
AQF – whether the model is specifically designed for air quality forecasting; 
Temp – models temporal dependencies; 
MultiV – supports multivariate predictions; 
Exog – incorporates future exogenous variables; 
Spat – models spatial correlations; 
Phy – integrates physical constraints.
}
\label{tab:methods}
\begin{tabular}{lcccccc}
\toprule
Methods             & AQF    & Temp   & MultiV & Exog   & Spat   & Phy   \\
\midrule
XGBoost             & \xmark & \xmark & \xmark & \cmark & \xmark & \xmark \\
LightGBM            & \xmark & \xmark & \xmark & \cmark & \xmark & \xmark \\
GC-LSTM             & \cmark & \cmark & \xmark & \xmark & \cmark & \xmark \\
\PM-GNN             & \cmark & \cmark & \xmark & \cmark & \cmark & \xmark \\
iTransformer        & \xmark & \cmark & \cmark & \xmark & \xmark & \xmark \\
TimeXer             & \xmark & \cmark & \cmark & \xmark & \xmark & \xmark \\
AirPhyNet           & \cmark & \cmark & \xmark & \xmark & \cmark & \cmark \\
\midrule
CMAQ                & \cmark & \cmark & \cmark & \cmark & \cmark & \cmark \\
\textbf{\OurModel}  & \cmark & \cmark & \cmark & \cmark & \cmark & \cmark \\
\bottomrule
\end{tabular}
\end{table}

\subsection{Evaluation Metrics}
\label{appendix:metrics}

To evaluate model performance, we use the following standard forecasting metrics:

\paragraph{Mean Absolute Error (MAE)}
\begin{equation}
\mathrm{MAE} = \frac{1}{N} \sum_{n=1}^{N} \left| \hat{\mathbf{X}}_n - \mathbf{X}_n \right|.
\end{equation}

\paragraph{Root Mean Square Error (RMSE)}
\begin{equation}
\mathrm{RMSE} = \sqrt{\frac{1}{N} \sum_{n=1}^{N} (\hat{\mathbf{X}}_n - \mathbf{X}_n)^2}.
\end{equation}

\subsection{Detailed Deployment and Case Studies}
\label{appendix:case_studies}

\subsubsection{Nationwide Air Quality During Festival}
\begin{figure}[h]
  \centering
  \includegraphics[width=\linewidth]{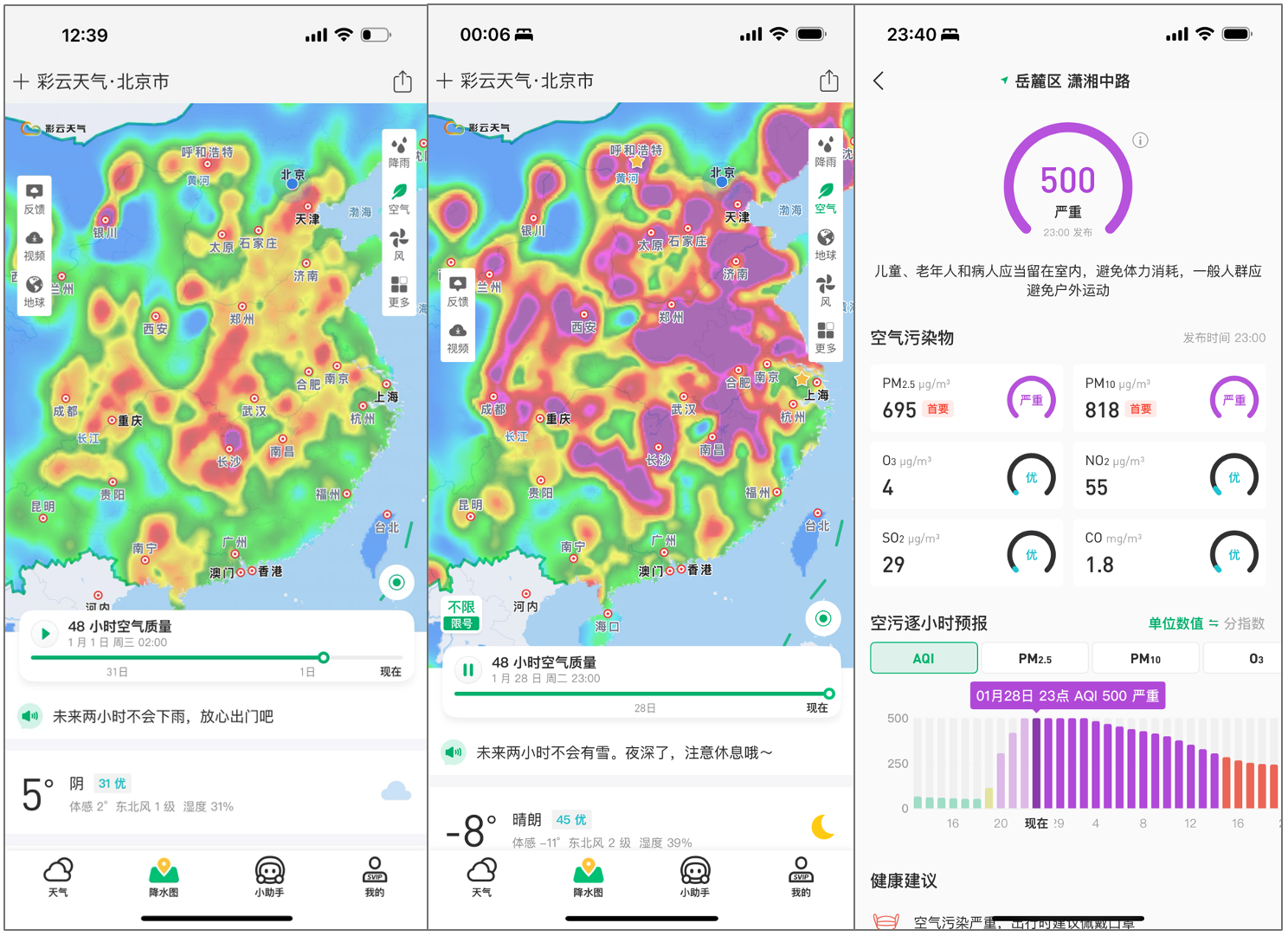}
  \caption{Nationwide air quality during New Year’s Eve and Spring Festival, highlighting severe pollution in Changsha caused by fireworks and its eventual dissipation forecasting.}
  \label{fig:nationwide_pollution}
\end{figure}

Figure~\ref{fig:nationwide_pollution} shows nationwide air quality distribution during the Chinese New Year and Spring Festival periods in 2025. The model successfully captured the severe pollution caused by extensive firework activities in regions like Changsha, predicting both the intensity of pollution and its eventual dissipation. This case highlights \OurModel's generalization capability to regions outside the training dataset and its ability to model short-term anthropogenic impacts.

\subsubsection{Extreme Pollution in Beijing}
\begin{figure}[h]
  \centering
  \includegraphics[width=\linewidth]{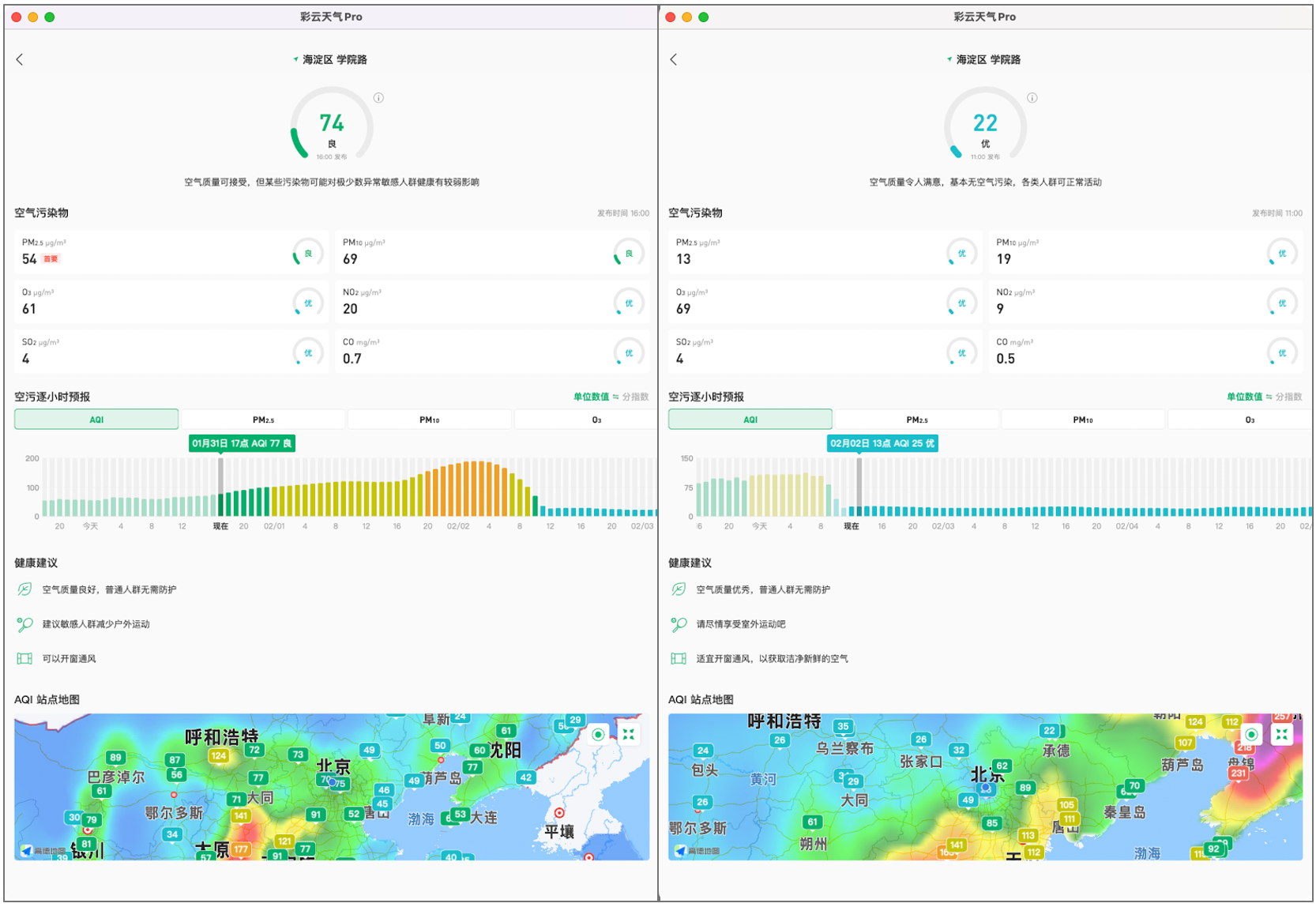}
  \caption{Model prediction (left) and real-world observation (right) for a pollution episode in Beijing, highlighting the system's early warning capability.}
  \label{fig:beijing_pollution}
\end{figure}

Figure~\ref{fig:beijing_pollution} illustrates a pollution episode in Beijing on January 31, 2025. The system forecasted the event 72 hours in advance, capturing the rapid \PM buildup due to high-pressure stagnation and local emissions. The prediction aligned closely with observed data, accurately signaling the event's onset and dissipation. This early warning capability underscores \OurModel's utility in proactive air quality management.

\subsubsection{California Wildfire Case}
\begin{figure}[h]
  \centering
  \includegraphics[width=\linewidth]{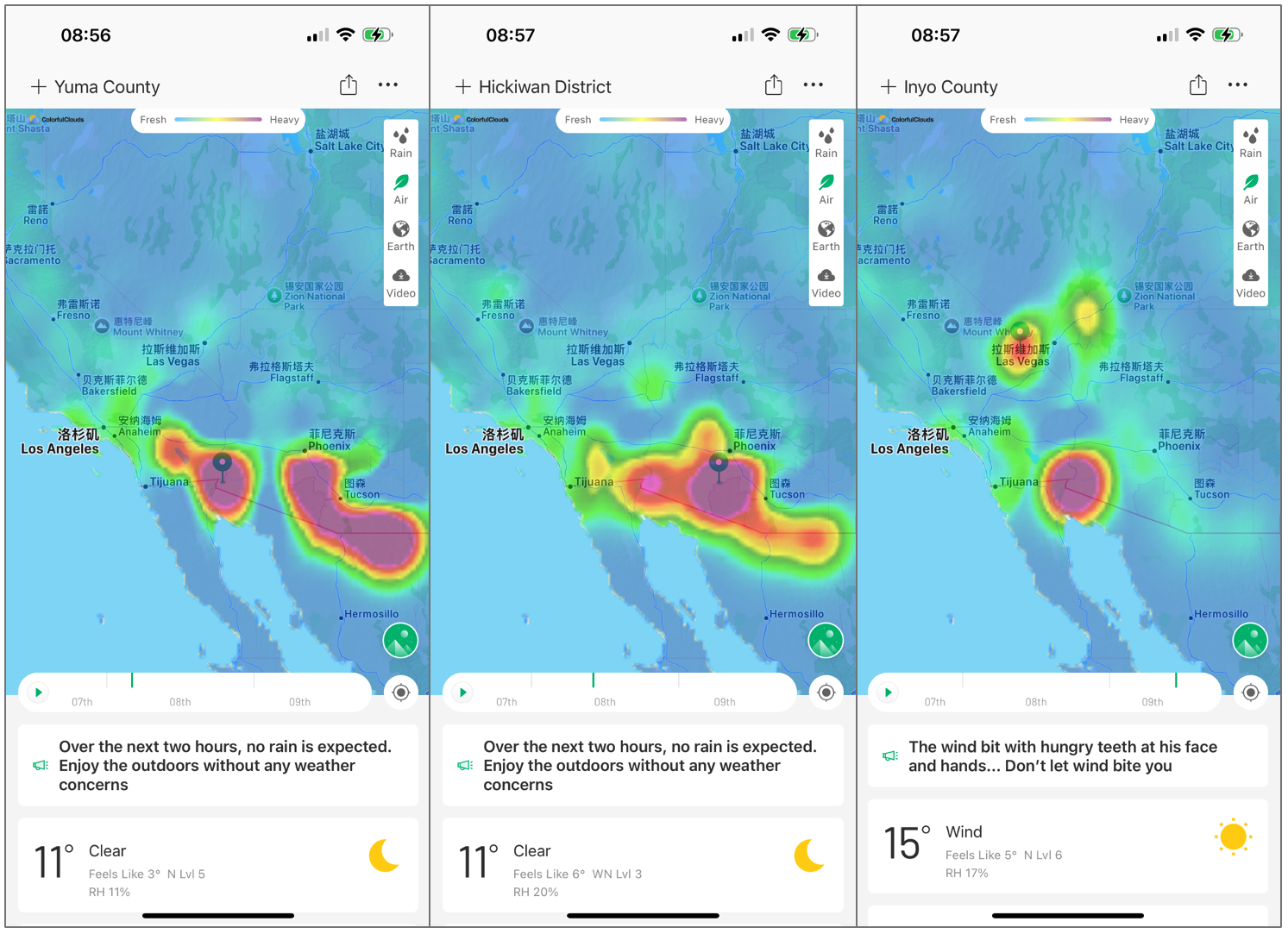}
  \caption{Air quality dynamics during California wildfires in January 2025, showcasing \OurModel's ability to track wildfire-induced pollution.}
  \label{fig:la_wildfire}
\end{figure}

In January 2025, wildfires in California caused severe air quality deterioration across the Los Angeles region. Figure~\ref{fig:la_wildfire} visualizes the event progression, with \OurModel\ capturing the pollutant spread driven by prevailing winds. The system's accurate tracking of wildfire-induced pollution demonstrates its versatility in addressing diverse atmospheric scenarios \cite{burke2023contribution} \footnote{\url{https://www.who.int/health-topics/wildfires}}.

\subsubsection{System Scalability and Performance}
\begin{figure}[h]
  \centering
  \includegraphics[width=\linewidth]{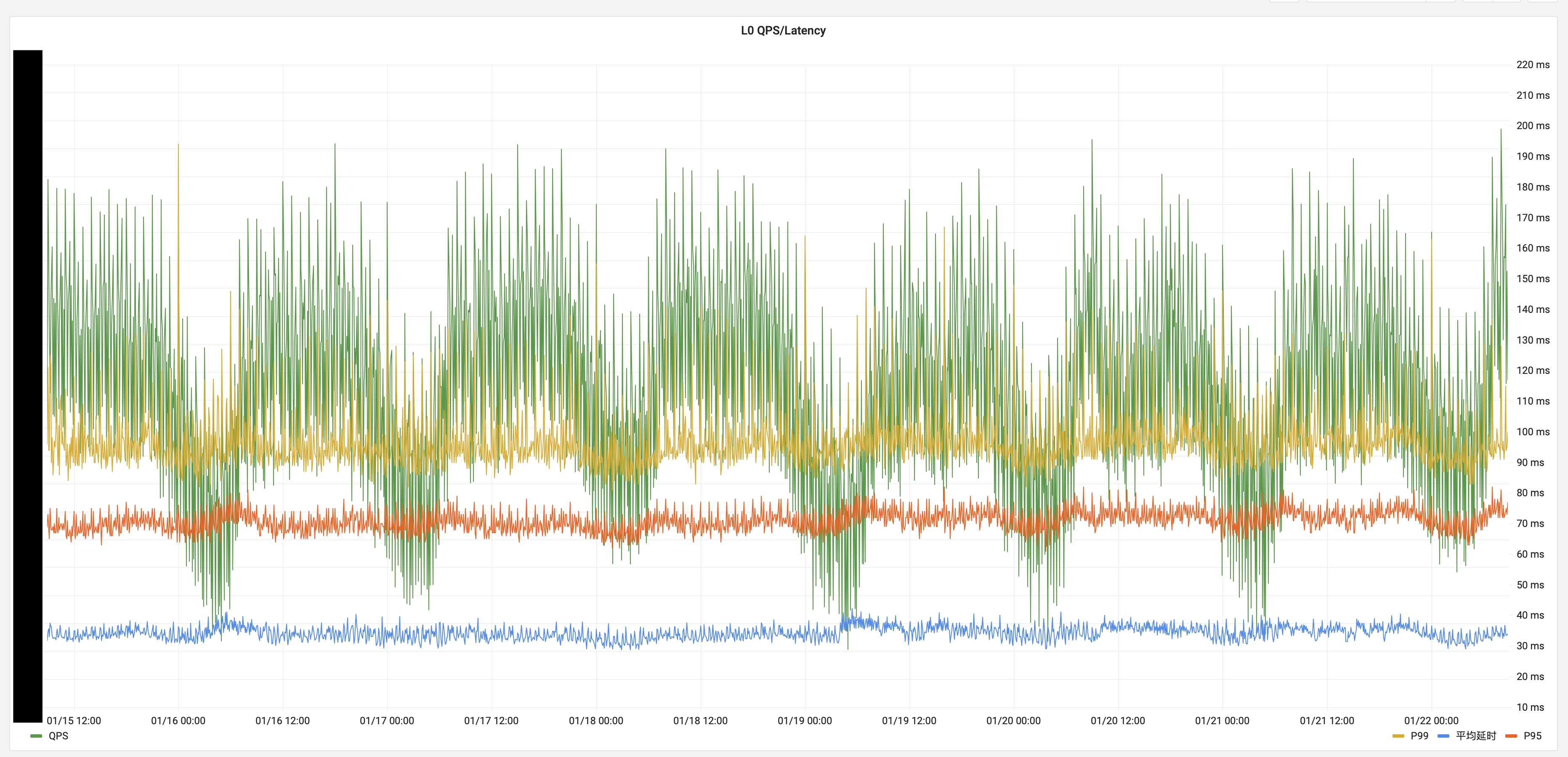}
  \caption{API performance metrics during deployment, showing low latency and consistent performance under high load conditions.}
  \label{fig:api_performance}
\end{figure}

To ensure robustness, the system was tested under high-demand conditions with concurrent API requests. Figure~\ref{fig:api_performance} showcases latency metrics, including P99, P95, and average latency, measured over a week. The results demonstrate the system's ability to handle high query-per-second (QPS) loads without performance degradation, affirming its scalability for public deployment.

\begin{figure}[h]
  \centering
  \includegraphics[width=1\linewidth]{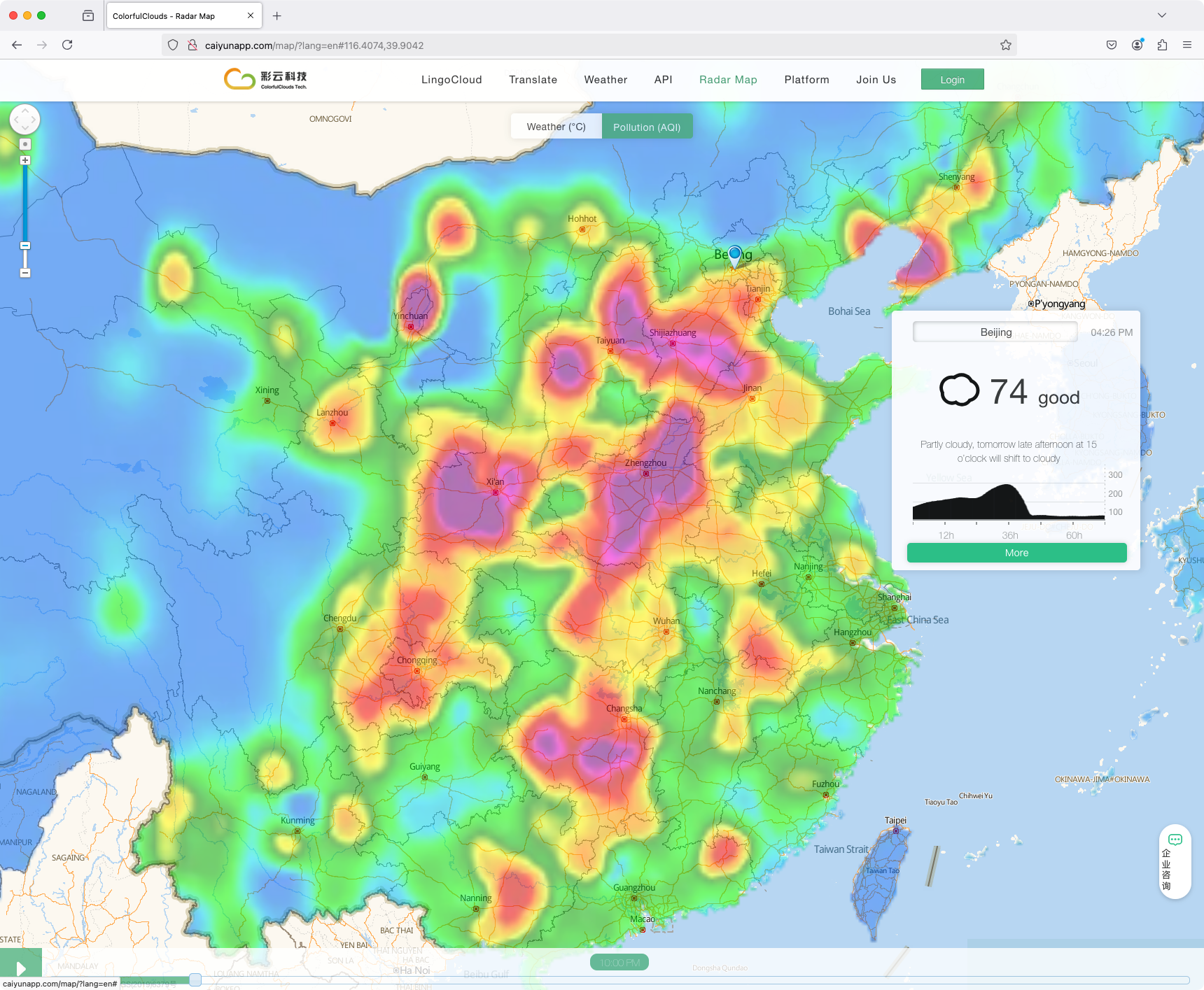}
  \caption{Real-time air quality visualization on the Caiyun web-based platform (\href{https://caiyunapp.com/map/\#116.3507,40.0099}{Website: https://caiyunapp.com/map}). The system provides 72-hour air quality forecasts, accessible without downloading an app, offering a seamless user experience directly via any browser.}
  \Description{}
  \label{fig:caiyun_web}
\end{figure}

\begin{figure}[h]
  \centering
  \includegraphics[width=0.8\linewidth]{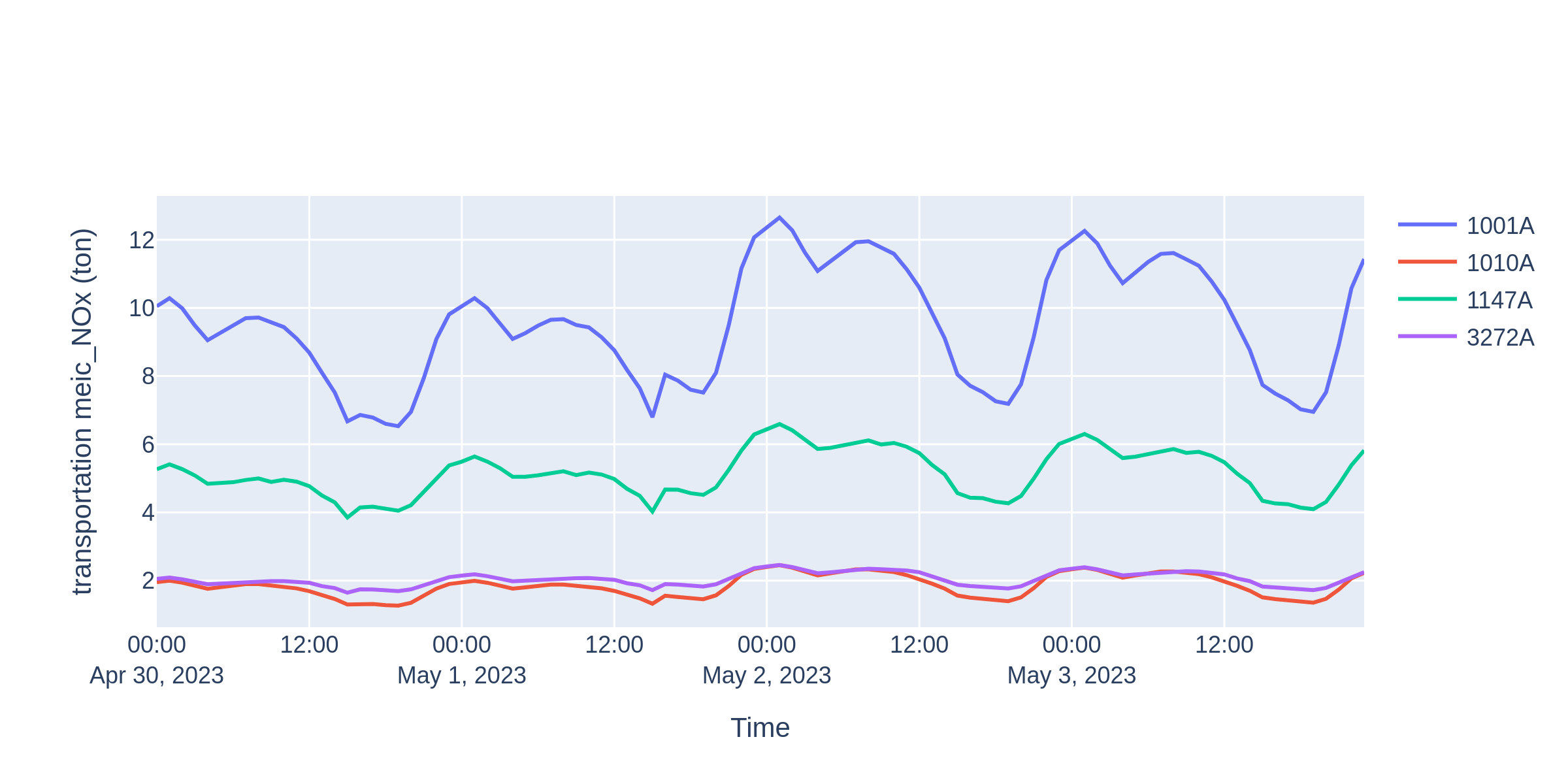}
  \caption{Temporal variation of \NitrogenOxides emissions at four air quality monitoring stations: 1001A (Beijing city center), 1010A (Beijing suburban), 1147A (Shanghai city center), and 3272A (Shanghai suburban). The data, converted to UTC+0, highlights distinct morning and evening peaks due to traffic-related \NitrogenOxides emissions. Urban stations exhibit higher levels compared to suburban stations.}
  \Description{}
  \label{fig:temporal_nox}
\end{figure}

\begin{figure}[h]
  \centering
  \includegraphics[width=\linewidth]{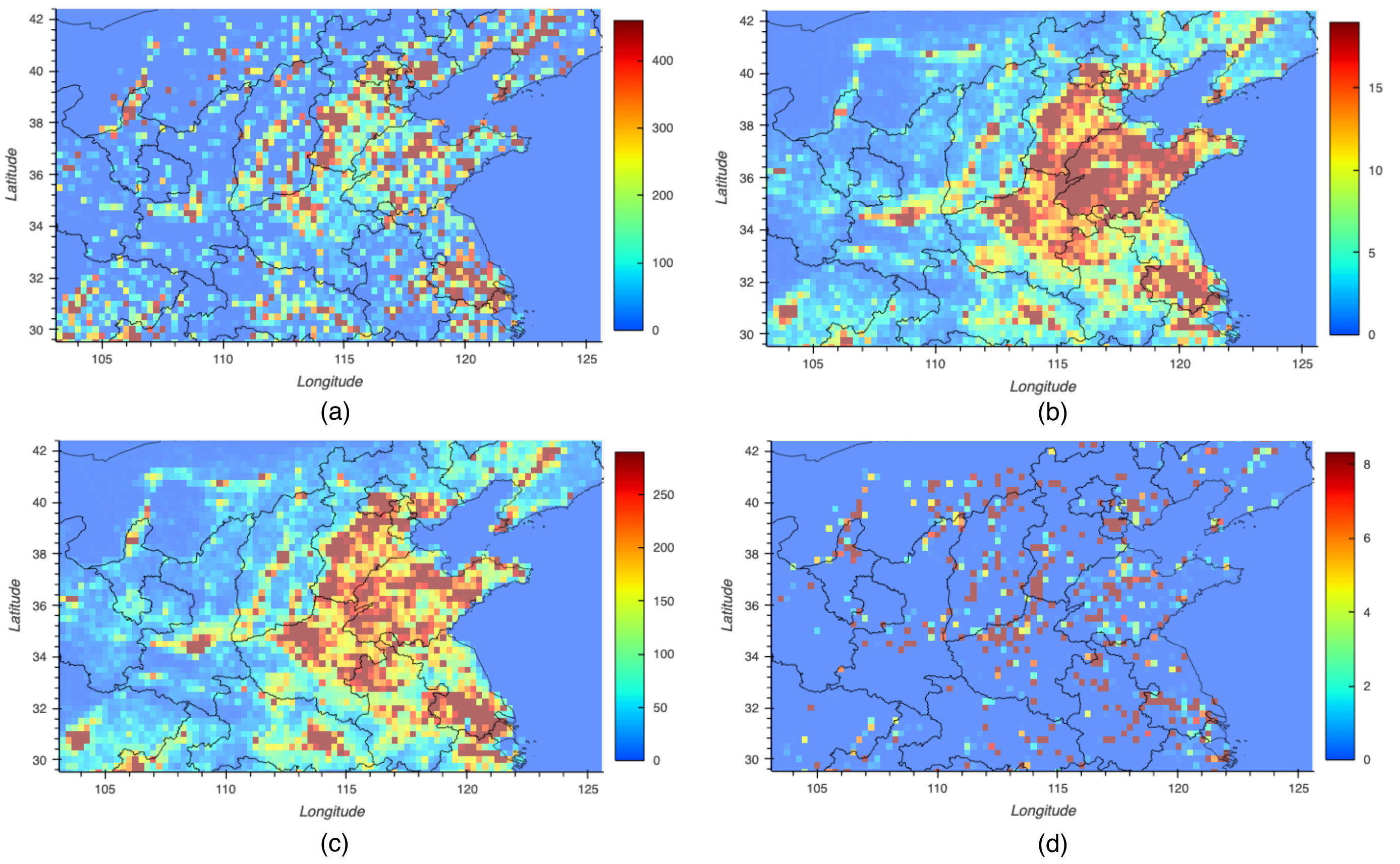}
  \caption{Spatial distribution of emissions over the study region:  
  (a) \NitrogenOxides emissions from the industrial sector (tons).  
  (b) Transport-related \PM emissions (tons).  
  (c) Transport-related \NitrogenOxides emissions (tons).  
  (d) Power sector \PM emissions (tons).  
  The maps showcase the spatial variability of emissions across different sectors, with industrial emissions and transport emissions concentrated in urban and industrial hubs.}
  \Description{}
  \label{fig:spatial_emissions}
\end{figure}

\subsubsection{Cross-Platform Accessibility}
\OurModel's deployment ensures easy access across multiple platforms. Users can download the \textbf{ColorfulClouds Weather} application from the Apple App Store or Android App Market for a comprehensive air quality forecasting experience. Alternatively, for those preferring a no-download option, the system is fully accessible through the \href{https://caiyunapp.com/map/\#116.3507,40.0099}{web-based interface: https://caiyunapp.com/map/}, offering real-time air quality predictions directly via any browser.

\subsection{Recognition in Competition}

\paragraph{Excellence in Air Quality Forecasting Competition}
The effectiveness of \OurModel has been further validated through its performance in real-world competitive scenarios. \OurModel achieved \textbf{first place in median performance} in the multi-city forecasting task of the prestigious \textbf{“Air Quality Forecasting in the Guangdong-Hong Kong-Macao Region 2024”} competition, hosted by the \textbf{China National Environmental Monitoring Center (CNEMC)}\footnote{National Ambient Air Quality Forecasting Model Comparison Platform: \url{http://124.128.14.106:10086/noticeDetail/66bef8dc65cfab60187f6887}}. 

This competition provided a rigorous benchmark for air quality forecasting models, requiring participants to produce accurate multi-day predictions across multiple cities in the Guangdong-Hong Kong-Macao region. \OurModel’s robust performance, particularly in capturing multi-day forecast trends, highlighted its advanced capabilities and domain-informed design, reinforcing its practical applicability and reliability in real-world forecasting tasks.

\subsection{Emissions Data Analysis}
\label{appendix:emissions}

The emissions data analyzed in this study include both temporal and spatial dimensions. The time series in Figure~\ref{fig:temporal_nox} demonstrates processed \NitrogenOxides emissions from the transportation sector at key air quality stations in Beijing (1001A, 1010A) and Shanghai (1147A, 3272A). The temporal trends reveal distinct peaks during morning and evening rush hours, aligning with traffic patterns. Emissions levels are notably higher at urban stations (1001A and 1147A) compared to suburban ones (1010A and 3272A), underscoring the significant contribution of transportation to \NitrogenOxides~emissions in urban areas.

The spatial distribution of emissions (Figure~\ref{fig:spatial_emissions}) provides insights into the geographic variability across sectors. Industrial \NitrogenOxides emissions (Figure~\ref{fig:spatial_emissions}a) are concentrated around major industrial zones. Transport-related \PM~and \NitrogenOxides~emissions (Figs.~\ref{fig:spatial_emissions}b and \ref{fig:spatial_emissions}c) are predominantly urban, reflecting the density of transportation networks in cities. Power sector \PM~emissions (Figure~\ref{fig:spatial_emissions}d) are distributed near energy generation facilities, indicating their localized impact. These spatial distributions emphasize the critical role of emissions from different sectors in shaping air quality across the region.

\begin{figure}[t]
  \centering
  \includegraphics[width=\linewidth]{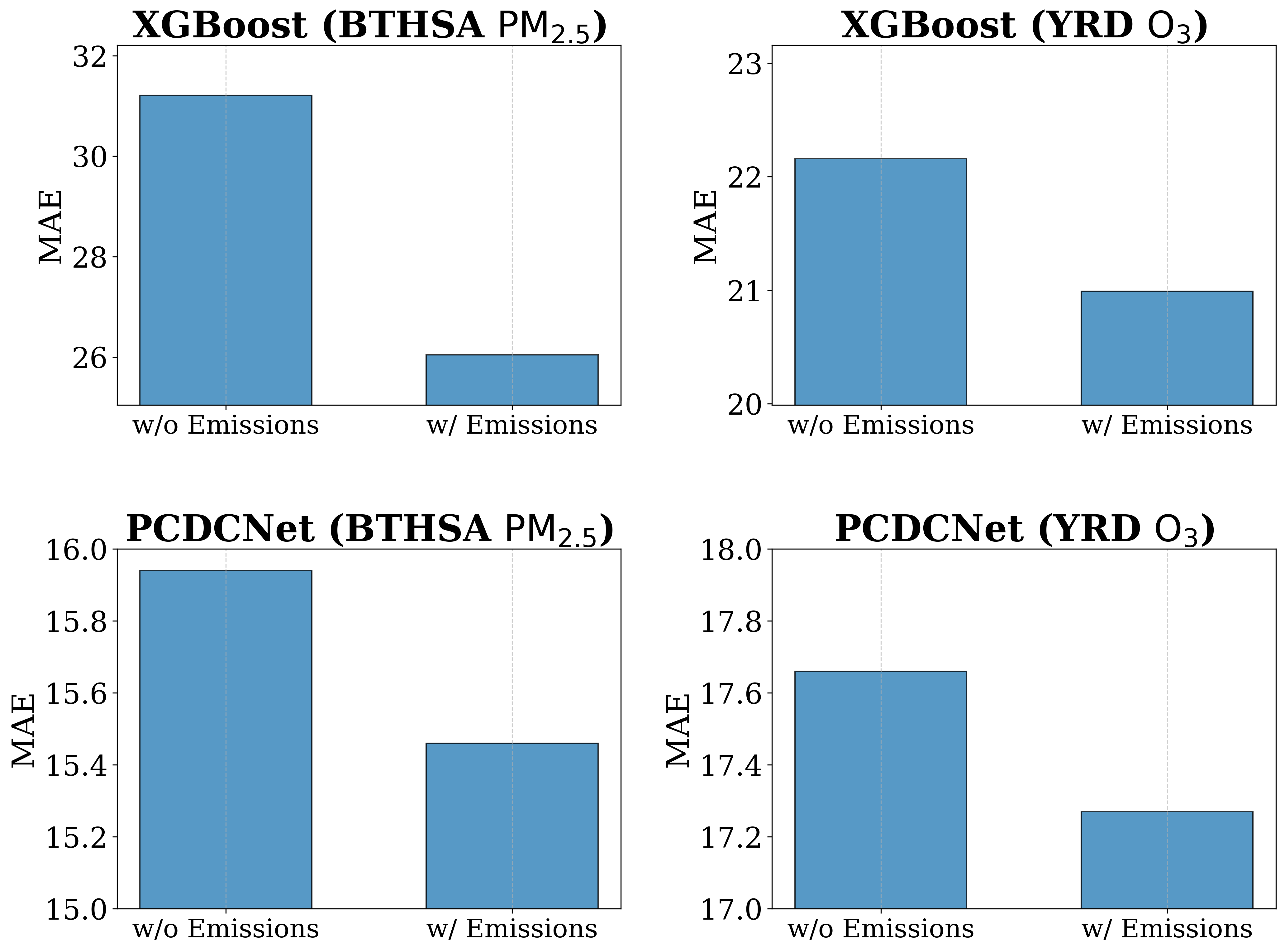}
    \caption{MAE comparison for BTHSA PM\(_{2.5}\) and YRD O\(_3\) predictions using XGBoost and PCDCNet, with and without emissions data.}
  \Description{}
    \label{fig:emissions_ablation}
\end{figure}

Figure~\ref{fig:emissions_ablation} illustrates the impact of emissions data on air quality predictions for BTHSA (\(\mathrm{PM}_{2.5}\)) and YRD (\(\mathrm{O}_3\)). The results demonstrate that incorporating emissions data significantly reduces MAE for both traditional models like XGBoost and advanced models like PCDCNet. This highlights the critical role of emissions data in capturing pollutant dynamics, particularly for chemically reactive pollutants like \(\mathrm{O}_3\). The consistent improvements across regions and models underscore the necessity of emissions data in accurate and robust air quality forecasting.

\subsection{Real-time Monitoring of Air Quality Forecast Accuracy}

\begin{figure*}[h]
  \centering
  \includegraphics[width=0.8\linewidth]{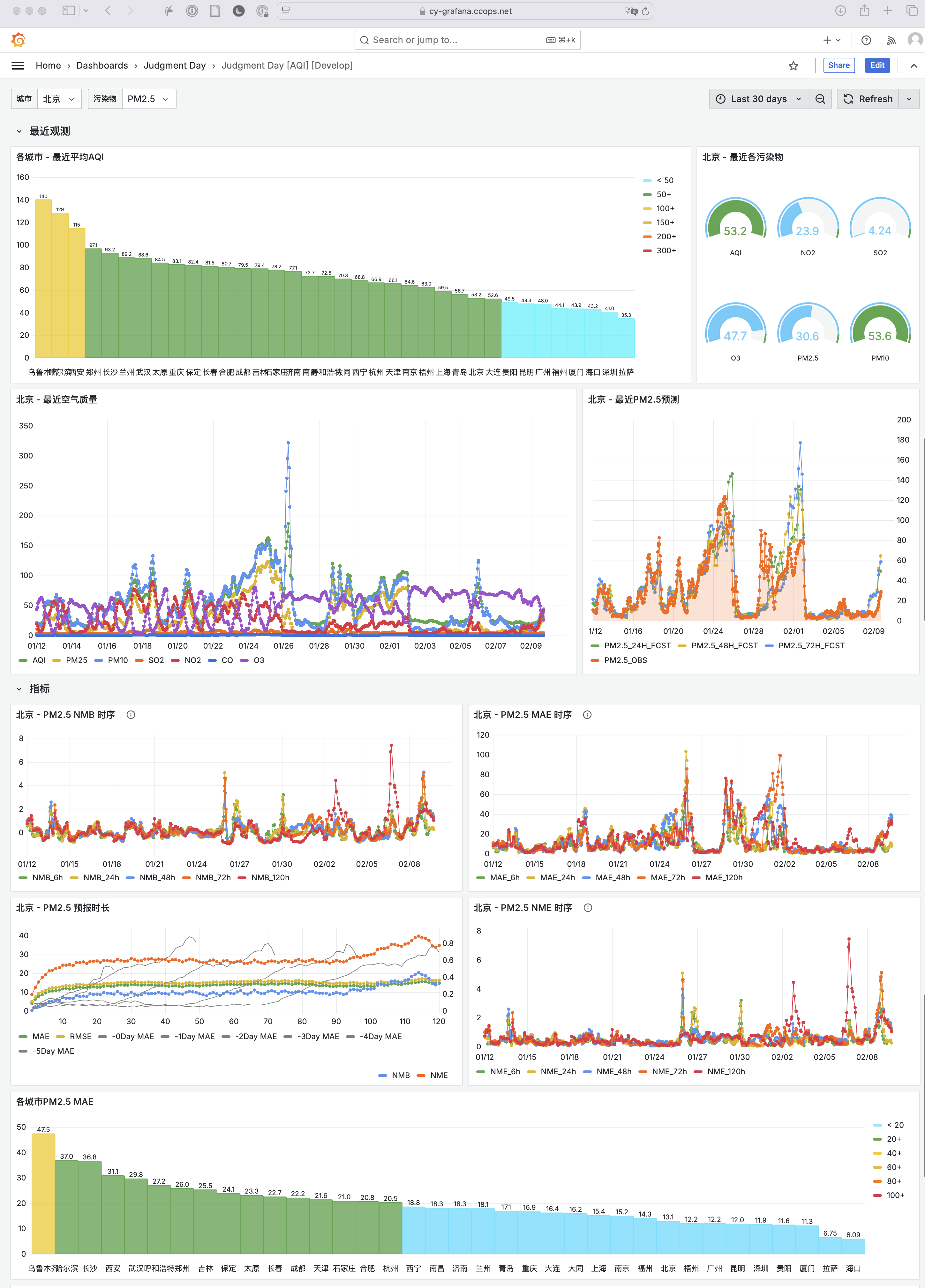}
  \caption{Online Real-Time Accuracy Monitoring of AQF}
  \label{fig:monitoring_accuracy}
\end{figure*}

The real-time monitoring dashboard in Figure~\ref{fig:monitoring_accuracy} presented above evaluates the accuracy of our air quality forecasting model. It provides key performance metrics for \PM predictions across different forecast horizons (e.g., 6 hours, 24 hours, and 72 hours) and various regions. The dashboard tracks metrics such as Mean Absolute Error (MAE), Normalized Mean Bias (NMB), and Normalized Mean Error (NME), which are essential for assessing the model's prediction accuracy. The bar charts on the right display the current air quality levels in Beijing, showing concentrations of pollutants like \PM, \PMcoarse, \ozone, and \SulfurDioxide. This monitoring system ensures that the forecasted air quality is continuously evaluated, providing real-time insights into model performance and facilitating further optimization of the forecasting system.

\end{document}